\documentclass[11pt]{article}
\usepackage[preprint]{acl}
\usepackage{times}
\usepackage{latexsym}
\usepackage[T1]{fontenc}
\usepackage[utf8]{inputenc}
\usepackage{microtype}
\usepackage{inconsolata}
\usepackage{graphicx}
\usepackage{booktabs}
\usepackage{multirow}
\usepackage{amsmath,amssymb}
\usepackage{xcolor}
\usepackage{tikz}
\usetikzlibrary{arrows.meta,positioning,fit}
\usepackage{pgfplots}
\pgfplotsset{compat=1.17}

\title{When Does Optical Context Compression Pay Off?\\A Decode-Aware Cost Analysis of Reasoning VLMs}

\author{Ziqing Ge \\
  Bitdeer Technologies Group \\
  \texttt{terrence.ge@bitdeer.com}}

\begin{document}
\maketitle

\begin{abstract}
Rendering long text as images lets a vision-language model (VLM) read a document with fewer input tokens, and evaluations often count mainly these prefill tokens. We show that for reasoning models this accounting is incomplete. Across five models and two long-context benchmarks, reading the rendered document makes the model \emph{think longer}: decode length grows by 1.5--3.5$\times$ on the four general-purpose VLMs tested, and under typical output pricing this decode tax makes image input \emph{more} expensive than text for two of the five models on QuALITY. Unrecoverable information loss is not sufficient to explain the tax at moderate compression: Qwen's OCR of the same images recovers 98.7\% of words, and feeding Kimi-K3 its own transcription as text restores text-mode reasoning length and accuracy. Whether compression pays off reduces to an \emph{exchange rate}, $(1{-}1/c)/(r{-}1)$, that must exceed the workload's price-weighted ratio of output to input tokens; this accounting decomposition shows why Kimi-K3 and DeepSeek-V4.1-Flash can benefit and Qwen and GLM cannot. None of the training-free signals we test predicts the tax before generation. As a practical heuristic, an image-first policy that sends only inputs with a per-input compression ratio of at least 1.6 to images and falls back to text once image-mode reasoning costs as much as the input's prefill saving cuts the bill by 6--14\% on Kimi-K3 and DeepSeek-V4.1-Flash at an accuracy cost of 0.3--1.2 points; on QuALITY, asking for a reasoning budget on both paths is a stronger lever. Code and data are available at \url{https://github.com/gitgzq/optical-compression-decode-tax}.
\end{abstract}

\section{Introduction}
Long contexts are expensive. A recent line of work proposes to compress them \emph{optically}: render the text into images and let a VLM read the pixels, so that a page of text costs several times fewer visual tokens than text tokens \citep{wei2025deepseekocr,cheng2025glyph}. The idea has drawn substantial follow-up work in 2026, covering long-context reasoning, selective expansion, adaptive resolution, code, tables, agent session history and latent reasoning \citep[e.g.,][]{wang2026vtcr,xie2026lensvlm,xu2026seer,zhong2026focusvtc,bhalgami2026pixels,alonso2026table,boluk2026snapcompact,wang2026renderofthought,lv2026onelatent}, and dedicated benchmarks \citep{zhao2025vtcbench}.

Much of this work measures cost mainly by the number of \emph{input} tokens, and some end-to-end comparisons fix the output length (256 tokens in \citealp{cheng2025glyph}). Others measure end-to-end latency with actual generation \citep{wang2026vtcr,zhong2026focusvtc,xu2026seer}, and SEER attributes its higher latency to longer completions (381 vs.\ 8 tokens); but these comparisons also differ in training or pipeline, which leaves the effect of the input modality itself on reasoning length and cost largely unisolated. For reasoning models, however, the dominant cost is often the \emph{output}: thinking traces of hundreds to thousands of tokens, which are also priced several times higher than input tokens by commercial APIs. If reading pixels changes how long a model thinks, the input savings may be partially or entirely consumed. Prior reports disagree on the direction: a practitioner report finds, on a SQuAD-based test, that three of four commercial reasoning models emit 2--7$\times$ more thinking tokens when the context is rendered as images (one rendering configuration, by our reading of its table) and calls the result a ``decode tax'' \citep{boluk2026snapcompact}, whereas a controlled study of open VLMs reports, for two of its models, outputs 5--19$\times$ \emph{shorter}, measured in characters, from images, which it calls a reasoning collapse \citep{sun2026reading}. One plausible difference, among others in models, tasks and rendering, is whether the model reasons explicitly: without a reasoning phase, harder retrieval may surface as premature, less accurate answers, while a thinking model may spend the difficulty as extra reasoning. Related observations: downsampled tables lead to longer and worse reasoning \citep{alonso2026table}; and selective-expansion pipelines report higher end-to-end latency than reading the full text \citep{xu2026seer}, or on short contexts \citep{xie2026lensvlm}. What is missing is a controlled study with a same-model text baseline, a characterization of how the tax scales, a diagnosis of where it comes from, and a criterion for when it makes optical compression a net loss.

We study this \emph{decode tax} on five VLMs---Qwen3.8-27B \citep{qwen2026qwen3827b}, GLM-5.3-Flash \citep{zaiorg2026glm53flash}, Kimi-K3 \citep{moonshotai2026kimik3}, DeepSeek-V4.1-Flash-1017 \citep{deepseek2026v41flash}, and Glyph \citep{cheng2025glyph}, a model trained specifically for visual text compression---on QuALITY \citep{pang2021quality} and LongBench \citep{bai2023longbench}, and, for the two models that can benefit, on the full Qasper, GovReport and QMSum test sets and RULER question answering at up to 128k tokens. Our contributions are:
\begin{enumerate}
  \item \textbf{A practical heuristic and its limits.} An image-first policy that falls back to text once image-mode reasoning exceeds a threshold set by the input's prefill saving uses the image path's own reasoning length as an online difficulty probe; gated on the per-input compression ratio, it saves 6--14\% on the two models that can benefit, with no learned component, and raises the LongBench saving from 4--10\% to 9--14\%. Reasoning budgets are the stronger lever on QuALITY: a mix of budgeted text and image paths matches or beats fallback (\S\ref{sec:fallback}).
  \item \textbf{Analysis.} An exchange-rate decomposition of the bill, the relative prefill saving per unit of relative decode increase against the workload's price-weighted output-to-input ratio, explains which models, renderings and workloads benefit; it is an accounting identity on the measured runs, not an out-of-sample predictor (\S\ref{sec:breakeven}). Unrecoverable information loss is not sufficient to explain the tax at moderate compression: Qwen's OCR of the same images recovers 98.7\% of words, and most of the extra reasoning occurs before the final answer is fixed (\S\ref{sec:mech}). Per-question routing has large oracle headroom (a per-question oracle saves 25--31\%), but none of the training-free predictors we test before generation recovers it, including a transport-efficiency router \citep{tang2026visual} applied with its published parameters (\S\ref{sec:route}).
  \item \textbf{Measurement.} With rendering parameters chosen per model for accuracy, image input saves 1.13--1.97$\times$ prefill but inflates decode by 1.5--3.5$\times$ on the four general-purpose VLMs tested; at list prices, Qwen and GLM pay more for images than for text, while Kimi-K3 and DeepSeek-V4.1-Flash pay 13--17\% less on QuALITY and 1.5--14\% less on LongBench (\S\ref{sec:tax}).
\end{enumerate}

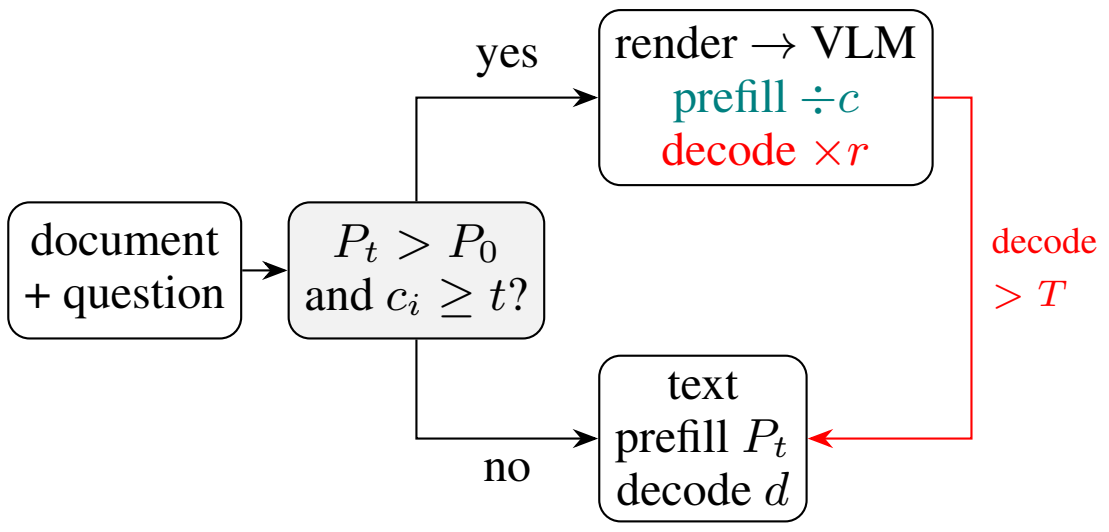
\begin{figure}[t]
\centering
\begin{tikzpicture}[font=\footnotesize, box/.style={draw, rounded corners, align=center, minimum height=0.9cm, inner sep=3pt}, >=Stealth]
\node[box] (doc) {document\\+ question};
\node[box, right=0.3cm of doc, fill=gray!10] (gate) {$P_t > P_0$\\and $c_i \ge t$?};
\node[box, above right=0.1cm and 0.35cm of gate] (img) {render $\to$ VLM\\\textcolor{teal}{prefill $\div c$}\\\textcolor{red}{decode $\times r$}};
\node[box, below right=0.1cm and 0.35cm of gate] (txt) {text\\prefill $P_t$\\decode $d$};
\draw[->] (doc) -- (gate);
\draw[->] (gate) |- node[pos=0.75, above]{yes} (img);
\draw[->] (gate) |- node[pos=0.75, below]{no} (txt);
\draw[->, red] (img.east) -- ++(0.25,0) |- node[pos=0.25, right, align=left]{\scriptsize decode\\\scriptsize $>T$} (txt.east);
\end{tikzpicture}
\caption{Optical compression divides prefill by the compression ratio $c$ but multiplies reasoning length by a decode multiplier $r$. Image-first fallback sends long, well-compressed inputs to the image path and aborts to text once image-mode reasoning exceeds $T$, the image-mode decode length whose output cost equals the input's prefill saving, $T=(P_t-P_v)/w$ (\S\ref{sec:fallback}); the aborted tokens are still paid. $P_t$, $P_v$: text and image prefill tokens; $c_i = P_t/P_v$: the input's own compression ratio; $d$: text-mode decode tokens; $w$: output-to-input price ratio; $P_0$, $t$: gate thresholds.}
\label{fig:overview}
\end{figure}

\section{Preliminaries}
\label{sec:prelim}
\subsection{Setup}
\label{sec:setup}
\paragraph{Models.} Qwen3.8-27B and GLM-5.3-Flash (Qwen and GLM below) are general reasoning VLMs; Kimi-K3 is a large MoE reasoning VLM served with tensor parallelism over 16 H200 GPUs; DeepSeek-V4.1-Flash-1017 (DeepSeek below) is a natively multimodal MoE (552B backbone parameters as stated in its model card) with its own ViT encoder, served as one 8-GPU replica per node; Glyph is GLM-4.1V-9B-Base continually trained for visual text compression \citep{cheng2025glyph}. Unlike Glyph, the other four are general-purpose VLMs without a disclosed training objective specifically targeting visual text compression. As a text-only reference for Glyph we also use GLM-4-9B-Chat-1M, following \citet{cheng2025glyph}.

\paragraph{Data.} QuALITY \citep{pang2021quality}: all 2{,}086 validation questions over 115 articles (multiple choice, $\sim$5.9k text tokens per article). LongBench \citep{bai2023longbench}: the 14 English, non-code subsets (2{,}750 examples; we report each subset's official metric, F1, ROUGE-L or accuracy, and for brevity call all of them accuracy); we exclude Chinese subsets (our renderer's font lacks CJK glyphs) and code subsets (the renderer normalizes whitespace, which destroys indentation). For Glyph we follow its official LongBench pipeline and rendering configuration. For Kimi-K3 and DeepSeek we add the complete test sets of three LongBench sources---Qasper \citep{dasigi2021qasper} (1{,}451 questions over 416 papers), GovReport \citep{huang2021govreport} (973 reports) and QMSum \citep{zhong2021qmsum} (281 queries over 35 meetings)---and RULER's single-hop QA task \citep{hsieh2024ruler} at 8k, 32k, 64k and 128k tokens (100 examples each); details in Appendix~\ref{app:details}.

\paragraph{Rendering.} Text is typeset in Times New Roman on a fixed-width canvas; a QuALITY article is one tall image and a LongBench document is paginated so that no page is downscaled (Appendix~\ref{app:details}). Because these general-purpose models degrade quickly with compression, we choose the font size and canvas width \emph{per model by accuracy} on QuALITY rather than reusing another model's working point, so QuALITY image accuracy is tuned on the test set, and LongBench, which did not enter this choice, is the cleaner test: Qwen uses 19\,px/1750\,px (1.36$\times$ prefill compression), GLM 18\,px or 16\,px (1.13$\times$/1.43$\times$), Kimi-K3 14\,px/1500\,px (1.78$\times$), DeepSeek 20\,px/1512\,px (1.97$\times$; pages cut between text lines to stay within its 1{,}024-token per-image limit, so no page is downscaled). %
We also render DeepSeek at 28\,px, where the images cost as many tokens as the text ($c{=}1.03$ on QuALITY, 0.93 on LongBench), as a no-compression control. %
At these working points the models' own transcriptions of the rendered QuALITY articles recover 98.7\% (Qwen) and 99.8\% (Kimi-K3) of words (\S\ref{sec:mech}, \S\ref{sec:diag}). %

\paragraph{Prompts and decoding.} On QuALITY the image arm uses the same prompt as the text arm with the document replaced by its rendered pages, placed before the question (for Glyph, following its official prompts, the document is replaced by ``The text is shown in the images''). Qwen and GLM use their recommended sampling (temperature 1.0) with 3 samples per question; Kimi-K3 uses greedy decoding plus 5 samples at temperature 0.8 on QuALITY and 3 samples at temperature 0.8 on LongBench; DeepSeek uses its recommended thinking mode (temperature 1.0, top-$p$ 1.0) with 3 samples on both benchmarks; Glyph uses its official decoding. On LongBench the text arm uses the official prompts and the image arm the Glyph image prompts, which replace the context by a pointer to the images but also reword some instructions (e.g., the multi-hop subsets drop the instruction repeated after the passages). The maximum generation length is 16k tokens (8k for Glyph, following its official setting) and truncated outputs count as wrong on QuALITY; on LongBench they are scored on whatever answer text they contain (Appendix~\ref{app:details}).

\paragraph{Cost.} For a request with $P$ prefill and $D$ decode tokens we report the bill $B = P + wD$, normalized by the all-text bill. For a question, $P_t$ and $P_v$ are the prefill tokens of its text and image inputs, $c_i = P_t/P_v$ is its compression ratio, and $D_t$ and $D_v$ are its text- and image-mode decode tokens (reasoning plus final answer). Over a set of questions, the compression ratio $c = \sum P_t/\sum P_v$ and the decode multiplier $r = \sum D_v/\sum D_t$ are ratios of totals (Table~\ref{tab:main}). We use $w{=}5$, matching common API price ratios between output and input tokens (and Kimi-K3's list price), and also report $w{=}1$ as an alternative price scenario and each model's own list-price ratio where it differs (GLM 3.3, DeepSeek 4, Qwen 6 at its international price; list prices at the time of our experiments, October--November 2026). The bill is a pricing model, not a compute model: it assumes constant per-token prices, ignoring long-context price tiers and prefix caching, whereas the compute and memory of each decode step grow with the KV cache, so a shorter image context can make decode tokens cheaper to serve under load (at concurrency 1 we measure no difference). We therefore report serving latency separately (\S\ref{sec:latency}).

\subsection{The Decode Tax}
\label{sec:tax}
\begin{table*}[t]
\centering
\small
\setlength{\tabcolsep}{4pt}
\resizebox{\textwidth}{!}{%
\begin{tabular}{llccccc}
\toprule
Model & Image setting & Acc.\ text & Acc.\ image & Prefill comp. & Decode img/text & Bill $w{=}5$ ($w{=}1$) \\
\midrule
Qwen & 19px/1750 & 92.65 & 88.22 & 1.36$\times$ & 2.14$\times$ & 1.256 (0.884) \\
Qwen & 14px/1500 & 92.65 & 80.90 & 2.41$\times$ & 3.48$\times$ & 1.554 (0.739) \\
GLM & 18px/1500 & 91.66 & 87.04 & 1.13$\times$ & 1.53$\times$ & 1.171 (0.973) \\
GLM & 16px/1500 & 91.66 & 83.24 & 1.43$\times$ & 2.12$\times$ & 1.334 (0.896) \\
Kimi-K3 (greedy) & 14px/1500 & 95.06 & 91.95 & 1.78$\times$ & 2.06$\times$ & 0.852 (0.631) \\
Kimi-K3 (temperature 0.8, 5 samples) & 14px/1500 & 94.73 & 92.51 & 1.78$\times$ & 2.07$\times$ & 0.871 (0.636) \\
DeepSeek (3 samples) & 20px/1512 & 93.85 & 91.24 & 1.97$\times$ & 2.47$\times$ & 0.898 (0.600) \\
DeepSeek (3 samples) & 28px/1512$^\ddagger$ & 93.85 & 92.91 & 1.03$\times$ & 1.55$\times$ & 1.090 (1.003) \\
Glyph & official & 79.29 & 68.07 & 1.79$\times$ & 1.13$\times$ & 0.787 (0.627) \\
\midrule
\multicolumn{7}{l}{\textit{LongBench (14 English subsets, 2{,}750 examples)}} \\
Kimi-K3 (temperature 0.8, 3 samples) & 14px/1500 & 57.11 & 56.43 & 1.77$\times$ & 2.97$\times$ & 0.985 (0.663) \\
DeepSeek (3 samples) & 20px/1512 & 57.65 & 56.19 & 1.84$\times$ & 2.59$\times$ & 0.929 (0.634) \\
DeepSeek (1 sample) & 28px/1512$^\ddagger$ & 57.31 & 56.27 & 0.93$\times$ & 2.26$\times$ & 1.302 (1.128) \\
Glyph & official & 49.79$^\dagger$ & 49.76$^\dagger$ & 2.6--2.9$\times^\dagger$ & -- & -- \\
\bottomrule
\end{tabular}}
\caption{Accuracy (\%; LongBench: macro-average of official subset metrics), prefill compression, decode inflation (ratio of total decode tokens) and normalized bill on QuALITY (2{,}086 questions) and LongBench (2{,}750 examples; 2{,}731 for DeepSeek, whose server limit excludes 19 \texttt{narrativeqa} examples, Appendix~\ref{app:details}). Bills below 1 mean the image path is cheaper than text. At DeepSeek's list price ($w{=}4$) its 20\,px bills are 0.833 (QuALITY) and 0.864 (LongBench). $^\dagger$Glyph on the official 21-subset LongBench with its own text baseline GLM-4-9B-Chat-1M as the text column; the two models differ, so decode ratios are not comparable there. $^\ddagger$No prefill compression; a control (\S\ref{sec:setup}). Bootstrap 95\% intervals for bills and accuracy changes are in Table~\ref{tab:ci}.}
\label{tab:main}
\end{table*}

Table~\ref{tab:main} summarizes the main measurements. Three observations stand out.

\paragraph{Prefill savings are modest once accuracy is protected.} General-purpose VLMs tolerate little compression: Qwen at 2.41$\times$ loses 11.8 points; at 1.36$\times$ it still loses 4.4. The working points that keep the accuracy gap within a few points compress prefill by only 1.13--1.97$\times$. For DeepSeek, the most compressive probed font (16\,px, 3.0$\times$) already loses 6.5 points on the probe subset. %

\paragraph{Decode inflates on every model.} Reading the image makes every model generate more tokens, by 2.1--3.5$\times$ for Qwen, 1.5--2.1$\times$ for GLM, 2.1--3.0$\times$ for Kimi-K3, 2.5--2.6$\times$ for DeepSeek, and 1.13$\times$ even for Glyph, which was trained on rendered text. The tax does not require compression: at 28\,px, where DeepSeek's images cost as many tokens as the text, decode still grows by 1.55--2.26$\times$. The inflation is not confined to failures: on questions answered correctly in both modes, the geometric-mean decode ratio is 2.5$\times$ for Qwen (19\,px), 1.45$\times$ for Kimi-K3 and 1.32$\times$ for Glyph. %

\paragraph{The net effect depends on the model.} With $w{=}5$, images are 17--55\% more expensive than text for Qwen and GLM, but 10--15\% cheaper for Kimi-K3 and DeepSeek on QuALITY (13--17\% at list prices, with $w{=}4$ for DeepSeek). At their own list-price ratios, GLM ($w{=}3.3$) pays 1.11 (18\,px) and 1.19 (16\,px) and Qwen ($w{=}6$) 1.32. Glyph is cheaper still, but at an 11-point accuracy loss relative to its own text mode. On LongBench the extra decode is almost entirely reasoning: final answers are as long as from text (0.96--1.03$\times$, including the summarization subsets). Kimi-K3's saving nearly disappears (bill 0.985): on \texttt{passage\_count}, reading the image drives decode up 9.4$\times$ and 26\% of image-mode responses hit the 16k limit inside the reasoning, which alone cancels most of what the other 13 subsets save (bill 0.783 without it). DeepSeek shows the same runaway on this subset ($r{=}12$). %

\begin{table}[t]
\centering
\small
\setlength{\tabcolsep}{3pt}
\resizebox{\columnwidth}{!}{%
\begin{tabular}{lcccccc}
\toprule
 & \multicolumn{3}{c}{Kimi-K3} & \multicolumn{3}{c}{DeepSeek} \\
\cmidrule(lr){2-4}\cmidrule(lr){5-7}
 & Score t / i & $r$ & Bill & Score t / i & $r$ & Bill \\
\midrule
Qasper & 60.1 / 60.4 & 2.04 & 0.889 & 58.5 / 57.8 & 1.66 & 0.796 \\
QMSum & 23.7 / 21.3 & 2.03 & 0.804 & 22.5 / 22.6 & 2.91 & 0.859 \\
GovReport & 33.6 / 33.5 & 1.43 & 1.052 & 36.3 / 34.0 & 1.59 & 1.091 \\
\midrule
RULER 8k & 97.7 / 92.7 & 1.37 & 0.682 & 95.3 / 92.3 & 1.84 & 0.690 \\
RULER 32k & 96.7 / 82.0 & 3.60 & 0.680 & 94.7 / 91.3 & 1.55 & 0.631 \\
RULER 64k & 95.3 / 82.0 & 5.75 & 0.681 & 94.0 / 89.0 & 2.09 & 0.627 \\
RULER 128k & 96.0 / 77.7 & 6.38 & 0.655 & 93.3 / 90.7 & 1.83 & 0.617 \\
\bottomrule
\end{tabular}}
\caption{Full-size test sets and RULER QA at increasing input length, 3 samples each: score text / image (official metrics), decode ratio $r$ and all-image bill at $w{=}5$ (DeepSeek at $w{=}4$: 0.767, 0.798, 1.040, 0.674, 0.627, 0.624, 0.616). Prefill compression: Kimi-K3 1.50 / 1.81 / 1.44 / 1.61, DeepSeek 1.59 / 1.94 / 1.50 / 1.64 (Qasper / QMSum / GovReport / RULER). Bootstrap 95\% intervals of the accuracy change: about $\pm$0.9 points (Qasper), $\pm$0.5 (QMSum, GovReport) and $\pm$3--7 (RULER, 100 questions per length); of the bill, $\pm$0.01--0.05.}
\label{tab:ext}
\end{table}

\paragraph{Full-size sets and longer inputs.} To check that the LongBench picture is not an artifact of its 200-example subsets, we ran Kimi-K3 and DeepSeek on the complete test sets of three of its sources and on RULER QA from 8k to 128k tokens, with the same rendering, prompts and sampling (Table~\ref{tab:ext}). Question answering over papers and meetings behaves as on LongBench: images are 11--20\% cheaper at an accuracy change of $-$2.4 to $+$0.3 points. Long-output summarization (GovReport) does not pay on either model (bill 1.05--1.09): text-mode outputs are already long (median 1.5--1.7k tokens), so a decode ratio of 1.4--1.6 outweighs the prefill saving. On RULER the two models diverge. DeepSeek stays within 2.6--5.0 points of text from 8k to 128k tokens at bills of 0.62--0.69, the largest all-image savings we measure, because answers are short and prefill dominates the cost. Kimi-K3 loses 5 points at 8k and 13--18 points from 32k on. Its image-mode failures are wrong answers, not truncated reasoning (zero-score rate 18--22\% vs.\ 3--5\% in text mode, at most 0.3\% truncated), and its decode ratio grows from 1.4 to 6.4 while the median ratio stays near 2, so the growth is in the tail. Whether a model can retrieve from many rendered pages is therefore a property of the model that the bill does not reveal: the accuracy side of the trade has to be checked per model at the target input length. %

\section{Where Does the Tax Go?}
\label{sec:mech}
\subsection{Readable but not read}
A natural explanation is that rendering loses information and the model must reason around the gaps. We test this directly on Qwen at its 1.36$\times$ working point by asking the same model to transcribe each rendered article. Aligned against the source, the transcription recovers 98.7\% of words (median 99.2\%). %
We then take the evidence sentences that the text-mode reasoning quotes and check how completely OCR recovers them. For questions answered correctly from text but incorrectly from the image, 93.3\% of evidence sentences are recovered at $\ge$95\% word recall, not significantly different from questions answered correctly in both modes (94.8\%; 196 of 210 vs.\ 4{,}134 of 4{,}359 sentences, Fisher's exact test $p{=}0.34$, which ignores the clustering of sentences within questions and so, if anything, overstates significance). At this working point the information is largely present in the visual tokens, so unrecoverable loss is not sufficient to explain the tax; this does not rule out partial loss (OCR misses 1.3\% of words) or a less reliable use of what is present. This echoes findings that VLMs can localize evidence in rendered text without utilizing it \citep{zeng2026magnifying} and the concurrent read--use gap \citep{shi2026aloud} and that self-reflection raises attention to the image far less than a user instruction does \citep{shi2026vlms}.

\subsection{The extra reasoning precedes the final answer}
To locate the extra tokens within a reasoning trace, we split each trace at two points: the first sentence that commits to a candidate option, and the point after which the model never changes its answer. This yields three phases: \emph{pre-commitment} (no candidate yet), \emph{oscillation} (candidates exist but change), and \emph{verification} (answer fixed). Commitments are detected with rules refined by manual inspection of traces. On question--sample pairs answered correctly in both modes, we attribute the extra image-mode decode to phases (Table~\ref{tab:phase}). %

\begin{table}[t]
\centering
\small
\setlength{\tabcolsep}{4pt}
\begin{tabular}{lrrrr}
\toprule
Model (setting) & Extra tok. & Pre-commit & Osc. & Verify \\
\midrule
Qwen 19px & +708 & 49\% & 34\% & 17\% \\
Qwen 14px & +1419 & 50\% & 34\% & 16\% \\
GLM 18px & +194 & 70\% & 12\% & 18\% \\
GLM 16px & +538 & 59\% & 25\% & 16\% \\
Glyph & +141 & 61\% & 13\% & 26\% \\
Kimi-K3 & +246 & 53\% & 27\% & 20\% \\
DeepSeek 20px & +486 & 59\% & 27\% & 14\% \\
\bottomrule
\end{tabular}
\caption{Share of the extra image-mode decode spent before any candidate answer (pre-commitment), while switching between candidates (oscillation), and after the final answer is fixed (verification). Pairs answered correctly in both modes with a detected commitment in both traces; QuALITY. The rule detects a commitment in 80--99\% of traces, except DeepSeek (59--69\%), whose terse traces often state the answer without a cue.}
\label{tab:phase}
\end{table}

Within this scope (QuALITY, pairs answered correctly in both modes), between 74\% and 86\% of the extra decode occurs \emph{before} the final answer is fixed, and, by the rule, the largest part before any answer is proposed; an LLM annotation confirms the first share and places the first commitment earlier (\S\ref{app:phasevalid}). Qwen additionally oscillates: in the 10\% of pairs with the largest inflation, it changes its answer 5.8 times on average when reading the image, versus 0.5 times when reading text. Post-commit verification---the target of early-exit methods---accounts for only 14--26\%. Kimi-K3 and DeepSeek, the two models that can profit from compression, follow the same pattern (53\% and 59\% before any candidate). Pre-commitment tokens may contain evidence search but also comparison and hedging; \S\ref{sec:diag} tests whether the extra reasoning is caused by the input modality or by content lost in rendering.

\subsection{How the inflation scales}
If the tax came from a fixed overhead (e.g., ``finding the page''), image-mode decode length $D_v$ would be the text-mode length $D_t$ plus a constant; if retrieval became uniformly harder, it would be a multiple of it. We compare $D_v{=}D_t{+}a$, $D_v{=}\hat r D_t$, $D_v{=}a{+}bD_t$ and $\log D_v{=}\log k{+}\beta\log D_t$ at the question level, with fitted constants $a$, $\hat r$, $b$, $k$ and $\beta$. To avoid correlated sampling noise, $D_t$ is averaged over one set of samples and $D_v$ over a disjoint set (Qwen and GLM: two text samples and the third image sample; Kimi-K3: two and three of its five samples); all four are fitted by least squares on $\log D_v$ (additive and affine constrained to positive predictions) on four fifths of the articles and evaluated on the held-out fifth by the median absolute log error and by the $R^2$ of $\log D_v$, the fitted loss (Table~\ref{tab:form}). %

\begin{table}[t]
\centering
\small
\setlength{\tabcolsep}{3pt}
\resizebox{\columnwidth}{!}{\begin{tabular}{lcccccc}
\toprule
 & \multicolumn{4}{c}{held-out median $|\Delta\log D_v|$} & & \\
\cmidrule(lr){2-5}
Model & add & mul & aff. & power & aff.\ $a$, $b$ & $\beta$ (corr.) \\
\midrule
Kimi-K3 & 0.40 & \textbf{0.39} & 0.40 & 0.41 & 46, 1.45 & 0.86 (1.17) \\
Qwen 19px & 0.53 & \textbf{0.51} & 0.53 & 0.52 & 347, 1.02 & 0.57 (0.65) \\
Qwen 14px & 0.79 & \textbf{0.68} & 0.78 & 0.75 & 632, 1.29 & 0.52 (0.59) \\
GLM 18px & 0.57 & 0.55 & 0.55 & \textbf{0.54} & 239, 0.58 & 0.60 (0.74) \\
GLM 16px & 0.72 & \textbf{0.68} & 0.70 & 0.71 & 312, 0.71 & 0.58 (0.73) \\
\bottomrule
\end{tabular}}
\caption{Functional form of decode inflation on QuALITY. Prediction error of image-mode reasoning length on held-out articles under additive ($D_t{+}a$), multiplicative ($\hat r D_t$), affine ($a{+}bD_t$, fitted intercept in tokens and slope, mean over folds) and power-law models, all fitted on $\log D_v$, and the power-law exponent $\beta$ (in parentheses: approximately noise-corrected, divided by the reliability of $\log D_t$, estimated from the Pearson correlation of log lengths between its two samples with a Spearman--Brown adjustment; this treats the log of the mean like a mean of logs, 95\% intervals by article resampling: Kimi-K3 0.80--0.92 and 1.09--1.26; all corrected values for Qwen and GLM below 0.82).}
\label{tab:form}
\end{table}

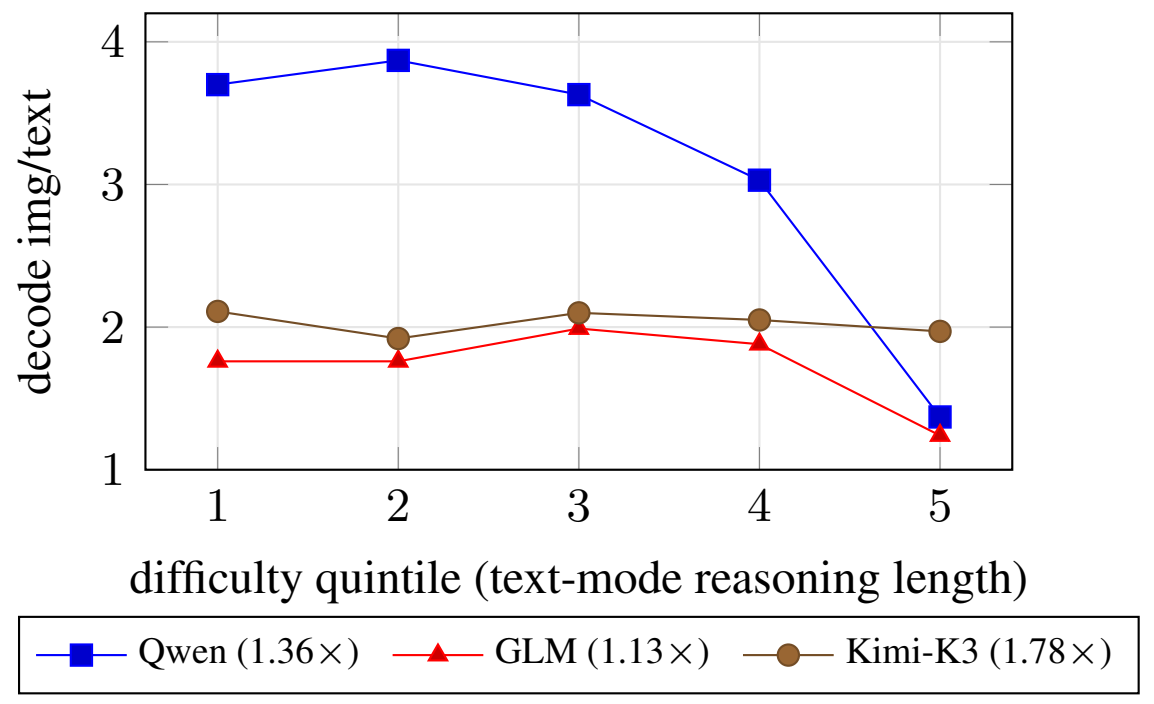
\begin{figure}[t]
\centering
\begin{tikzpicture}
\begin{axis}[width=0.95\columnwidth, height=4.6cm, xlabel={difficulty quintile (text-mode reasoning length)}, ylabel={decode img/text}, xtick={1,2,3,4,5}, ymin=1, ymax=4.2, legend style={font=\scriptsize, at={(0.5,-0.32)}, anchor=north, legend columns=3, /tikz/every even column/.append style={column sep=4pt}}, font=\footnotesize, grid=major, grid style={gray!20}]
\addplot+[mark=square*] table[x=q, y=qwen, col sep=comma] {data/quintile_decode_ratio.csv}; \addlegendentry{Qwen (1.36$\times$)}
\addplot+[mark=triangle*] table[x=q, y=glm, col sep=comma] {data/quintile_decode_ratio.csv}; \addlegendentry{GLM (1.13$\times$)}
\addplot+[mark=*] table[x=q, y=k3, col sep=comma] {data/quintile_decode_ratio.csv}; \addlegendentry{Kimi-K3 (1.78$\times$)}
\end{axis}
\end{tikzpicture}
\caption{Decode multiplier by difficulty quintile on QuALITY, with binning and measurement on disjoint samples to avoid regression to the mean. Flat for Kimi-K3 ($\beta{\approx}1$); declining for Qwen and GLM ($\beta{\approx}0.6$--$0.75$).}
\label{fig:quintile}
\end{figure}

Question-level prediction error does not separate the forms: fitted on the same loss, all four are within 0.04 of each other, except for Qwen at 14\,px, where the multiplicative model is better by 0.07--0.11 (fits on the raw length scale, which the long tail dominates, are 0.3--0.6 worse). On the fitted loss the power law is best in every setting, narrowly for Kimi-K3 (held-out $R^2$ 0.39 vs.\ 0.38 multiplicative) and clearly for Qwen and GLM (0.18--0.31 vs.\ 0.03--0.14), consistent with their sub-linear inflation below. The shape shows instead in the exponent and the quintile ratios. For Kimi-K3 the exponent is 0.86 (95\% interval 0.80--0.92, articles resampled) before and 1.17 (1.09--1.26) after the approximate noise correction, so the two bracket 1 and the correction dominates the uncertainty; the affine intercept is small (46 tokens), and the multiplier is flat across difficulty quintiles (1.9--2.1$\times$, Figure~\ref{fig:quintile}): reading pixels lengthens reasoning by a roughly constant factor. For Qwen and GLM the corrected exponent is 0.59--0.74, with every interval below 0.82, and the multiplier shrinks with difficulty (Qwen: 3.7$\times$ in the easiest quintile, 1.4$\times$ in the hardest): their inflation is sub-linear in text-mode length, as a fixed overhead plus a proportional term would produce (affine intercepts of 240--630 tokens). The decline is not an artifact of the generation limit: excluding every question truncated in either mode leaves the hardest-quintile ratios unchanged. Binning and measuring on the same samples would add a further artifact, regression to the mean, which inflates the ratio in the easiest bin; Figure~\ref{fig:quintile} bins and measures on disjoint samples. Accuracy losses grow with difficulty (Kimi-K3: $-$0.6 points in the easiest quintile, $-$3.4 in the hardest). %
On LongBench, where reasoning lengths differ by two orders of magnitude across tasks, Kimi-K3's exponent is again close to 1 ($\beta{=}0.99$ on non-truncated examples), and leave-one-subset-out errors again do not separate the forms, with the additive model, if anything, lowest (additive 0.43, multiplicative 0.47, affine 0.48, power 0.48), so the exponent there does not by itself single out a multiplicative form; DeepSeek has $\beta{=}0.82$ on QuALITY and 1.01 on LongBench (not noise-corrected, from means over the same samples). %
Note that the aggregate multiplier hides opposite shapes: Kimi-K3, Qwen (19\,px) and GLM (16\,px) all have a total decode ratio of 2.1, but Kimi-K3's multiplier is flat across quintiles (Figure~\ref{fig:quintile}), while Qwen's falls from 3.7$\times$ on the easiest quintile to 1.4$\times$ on the hardest and GLM's from 2.9$\times$ to 1.6$\times$ (Figure~\ref{fig:quintile} shows GLM at 18\,px, which falls from 1.8$\times$ to 1.2$\times$), so their tax is largest exactly on the easy questions that an image path should serve. %

\subsection{A heavy tail of runaway reasoning}
\label{sec:tail}
Beyond the bulk, image-mode decode has a heavy tail. On QuALITY the top 10\% of image-mode responses account for 51--61\% of all image-mode decode tokens on Kimi-K3, Qwen and GLM, and the tail is present even in the easiest quintile (GLM at 16\,px: mean 682 vs.\ median 300 tokens). The tail is only partly a property of the question: a response above the 90th percentile is followed by another one on the next sample of the same question 43--58\% of the time (10\% by chance). At the extreme, on LongBench \texttt{passage\_count}, a quarter of Kimi-K3's image-mode responses reason until the 16k limit without producing an answer; since no text-mode response on this subset reaches the limit, $r$ and the normalized bill there are lower bounds (truncation rates in Appendix~\ref{app:details}). The empirical mean residual life of an unfinished image-mode response \emph{grows} with its current length (DeepSeek on QuALITY: 1.2k remaining tokens at 256, 3.5k at 8k), consistent with a long tail over the observed range: the longer a response has run, the longer it is expected to keep running. A single runaway costs the savings of several ordinary questions, which is why both all-image bills and any routing method are dominated by how they treat the tail. %

\subsection{Information loss likely adds on top at high compression}
The picture changes at higher compression. Lowering Qwen's compression from 2.41$\times$ toward its 1.36$\times$ working point shortens the decode tail substantially (90th percentile 6{,}861$\rightarrow$3{,}543 tokens; outputs over 3k tokens 24.3\%$\rightarrow$12.0\%) while raising accuracy by 2.5--7.3 points. %
OCR fidelity does degrade at somewhat higher compression (90\% word recall at $\approx$2.7--3.1$\times$ for Qwen and Kimi-K3), so missing information plausibly compounds the retrieval problem at 2.41$\times$, although we did not measure OCR recall at that setting.

\subsection{Diagnostic interventions}
\label{sec:diag}
High OCR recovery shows that the information \emph{can} be extracted, not that the model extracts the decisive negations, references and qualifiers when answering. We test this with an \emph{OCR round-trip}: the model's own transcription of the rendered document is fed back as text input. If reasoning length returns to the text-mode level, lost content is not needed to explain the tax; if it stays high, transcription errors are implicated.

We run the round-trip on Kimi-K3 over all 115 QuALITY articles. The model transcribes each rendered article with greedy decoding (one transcription that hit the length limit was re-run at temperature 0.8); mean word recall is 99.8\% (minimum 97.0\%). The transcript then replaces the article in the text prompt, everything else unchanged, with the first three of the five samples of Table~\ref{tab:main} (samples 0--2, 6{,}258 paired answers). The round-trip arm behaves like text: accuracy is 94.66\% against 94.79\% for text and 92.41\% for image on the same three samples, and decode length is 0.97$\times$ text against 2.09$\times$ for image. The same holds for the 159 answers on articles transcribed below 98\% recall (0.81$\times$ vs.\ 1.60$\times$). Given the same content as text, even text it wrote itself from the image, the model reasons as briefly as on the original: unrecoverable content loss is not sufficient to explain the tax in this setting, consistent with the read--use gap reported concurrently for accuracy \citep{shi2026aloud}; what the round trip adds is that decode length, not only accuracy, returns to the text level. As a pipeline, transcribing the whole document does not pay: with thinking enabled it costs 12.3k decode tokens per article, about the bill of nine text questions, so even shared across an article's 18 questions and three seeds the round-trip bill is 1.12 (0.95 without the transcription, against 0.87 for direct image input). Transcribing only the relevant pages in a separate call, as in concurrent work \citep{shi2026aloud}, would cost less; we did not test that pipeline. The variant we did test, locating and transcribing the relevant passage within the same response, does not help hard questions (\S\ref{sec:fail}). %

\section{When Does Optical Compression Pay Off?}
\label{sec:breakeven}
\label{sec:analysis}

For a single question, write $P = P_t$ and $d = D_t$, and let $c$ and $r$ denote its own ratios $P_t/P_v$ and $D_v/D_t$. The image path costs $P/c + w\,r\,d$ and the text path $P + w\,d$, so for $r>1$ the image path is cheaper iff
\begin{equation}
d < d^\ast = \frac{(1-1/c)\,P}{w\,(r-1)}.
\label{eq:dstar}
\end{equation}
(for $r\le1$ it is cheaper whenever $c>1$). This is the modality analogue of the break-even analysis for prompt compression in \citet{johnson2026prompt}. When inflation follows the power law $D_v = k\,d^{\beta}$, the condition generalizes to $w\,(k\,d^{\beta}-d) < (1-1/c)\,P$; Eq.~\ref{eq:dstar} is the case $\beta{=}1$, $k{=}r$, which describes Kimi-K3. In either form, $P$ and $c$ are known from the input before generation, but $d$ and the question's own $r$ are not: a router must predict $d$ and rely on an $r$ (or $k$, $\beta$) calibrated on other questions, so the per-question net saving $S = P(1-1/c) - w\,(D_v - d)$ is the natural routing target. On QuALITY, where documents have similar length, $d$ alone is an adequate proxy; on LongBench, where $P$ varies severalfold, the threshold must scale with $P$.

\begin{table}[t]
\centering
\small
\setlength{\tabcolsep}{4pt}
\begin{tabular}{lcccc}
\toprule
Model & $c$ & $r$ & $d^\ast$ & $d$ (easiest 20\%) \\
\midrule
Kimi-K3 & 1.78 & 2.11 & $\approx$465 & 132 \\
Qwen & 1.36 & 3.70 & $\approx$117 & 133 \\
GLM & 1.13 & 1.76 & $\approx$180 & 286 \\
\bottomrule
\end{tabular}
\caption{Break-even reasoning length ($w{=}5$, $P\approx$5.9k; $r$ is the decode multiplier on the easiest quintile, Figure~\ref{fig:quintile}) versus the typical text-mode reasoning length of the easiest quintile on QuALITY. Routing can only help when a substantial fraction of questions falls below $d^\ast$.}
\label{tab:breakeven}
\end{table}

Table~\ref{tab:breakeven} is consistent with this condition. For Kimi-K3, $d^\ast\approx465$ exceeds the reasoning length of the first four quintiles, and indeed the image path is cheaper in those quintiles (bill ratio 0.72--0.84) and more expensive only in the hardest one (1.09). For Qwen, even the easiest quintile sits at $d^\ast$, and for GLM above it; consistently, the image path is more expensive in every quintile for both. A model benefits from optical compression only if it combines a high tolerable compression ratio and a low decode multiplier with text-mode reasoning that is short relative to the input.

\paragraph{The workload form: an exchange rate.} Summed over a workload with aggregate ratios $c = \sum P_t/\sum P_v$ and $r = \sum D_v/\sum d$, and with $\bar P_t$ and $\bar d$ the mean text prefill and text-mode decode tokens per question, the image path is cheaper, for $r>1$, iff
\begin{equation}
E \equiv \frac{1-1/c}{r-1} \;>\; \frac{w\,\bar d}{\bar P_t}.
\label{eq:rate}
\end{equation}
The left side is an exchange rate: the relative prefill saving per unit of relative decode increase (in absolute terms, $E\,\bar P_t/\bar d$ input tokens saved per extra output token). It depends on the model and the rendering but also on the workload, through $r$ (Kimi-K3: 0.41 on QuALITY, 0.22 on LongBench). The right side depends on the workload and the price: the price-weighted ratio of text-mode output to input. Because $c$ and $r$ are measured on the same runs, the equivalence with a bill below 1 is an exact accounting identity, not an out-of-sample prediction; its use is the decomposition (Table~\ref{tab:rate}). %

\begin{table}[t]
\centering
\small
\setlength{\tabcolsep}{3.5pt}
\resizebox{\columnwidth}{!}{%
\begin{tabular}{llccccc}
\toprule
Model & Data & $c$ & $r$ & $E$ & $w\bar d/\bar P_t$ & Cheaper? \\
\midrule
Kimi-K3 & QuALITY & 1.78 & 2.07 & \textbf{0.41} & 0.26 & yes \\
Kimi-K3 & LongBench & 1.77 & 2.97 & \textbf{0.22} & 0.21 & yes \\
DeepSeek 20px & QuALITY & 1.97 & 2.47 & \textbf{0.34} & 0.20 & yes \\
DeepSeek 20px & LongBench & 1.84 & 2.59 & \textbf{0.29} & 0.19 & yes \\
DeepSeek 28px & QuALITY & 1.03 & 1.55 & 0.05 & 0.20 & no \\
Qwen 19px & QuALITY & 1.36 & 2.14 & 0.23 & 0.59 & no \\
GLM 16px & QuALITY & 1.43 & 2.12 & 0.27 & 0.81 & no \\
GLM 18px & QuALITY & 1.13 & 1.53 & 0.22 & 0.81 & no \\
Glyph & QuALITY & 1.79 & 1.13 & \textbf{3.45} & 0.67 & yes \\
\bottomrule
\end{tabular}}
\caption{Exchange rate $E$ (Eq.~\ref{eq:rate}) versus the workload's price-weighted output-to-input ratio at $w{=}5$, except DeepSeek at its list price $w{=}4$. Glyph is cheaper but loses 11.2 points (Table~\ref{tab:main}). Bold and \emph{yes}: $E$ exceeds the right side, i.e., all-image is cheaper than text.}
\label{tab:rate}
\end{table}

Three things follow. First, Qwen and GLM fail mainly because they reason 2--3$\times$ longer in text mode (mean 702 and 928 tokens vs.\ 302), which raises the bar their exchange rate must clear 2.3--3.1-fold; on the same workload their exchange rate is also lower (0.22--0.27 vs.\ 0.41 for Kimi-K3), though comparable to Kimi-K3's on LongBench. Second, compression cannot be bought by rendering smaller: $r$ grows with $c$ (Qwen: $c$ 1.24$\rightarrow$2.41, $r$ 2.09$\rightarrow$3.48, $E$ only 0.18$\rightarrow$0.24), and at the other end DeepSeek's 28\,px rendering has one of the lowest $r$ but $E\approx0$. Third, the same model can sit on either side depending on the workload: Kimi-K3 clears the bar by a wide margin on QuALITY ($E{=}0.41$ vs.\ 0.26) but only barely on LongBench (0.22 vs.\ 0.21), where runaway reasoning on one subset pushes $r$ from 1.9 without it to 3.0, and on the full GovReport set its all-image bill is above 1 (1.052, Table~\ref{tab:ext}). The per-question form of the same accounting, $P_t - P_v > w(D_v - d)$, sets the fallback threshold in \S\ref{sec:fallback}. %

\section{Can the Tax Be Avoided per Question?}
\label{sec:route}
\paragraph{Setup.} For a question with text prefill $P_t$, image prefill $P_v$, text-mode reasoning length $d$ and image-mode length $D_v$, the image path saves $S = P_t - P_v - w\,(D_v - d)$. $P_t$ and $P_v$ are known before generation (the number of visual tokens is fixed by the page count); $d$ and $D_v$ are not. A router sends a question to the image path when a predicted saving or a predicted difficulty is favorable, and any cost of producing the prediction is charged to the routed bill. Accuracy budgets are relative to all-text.

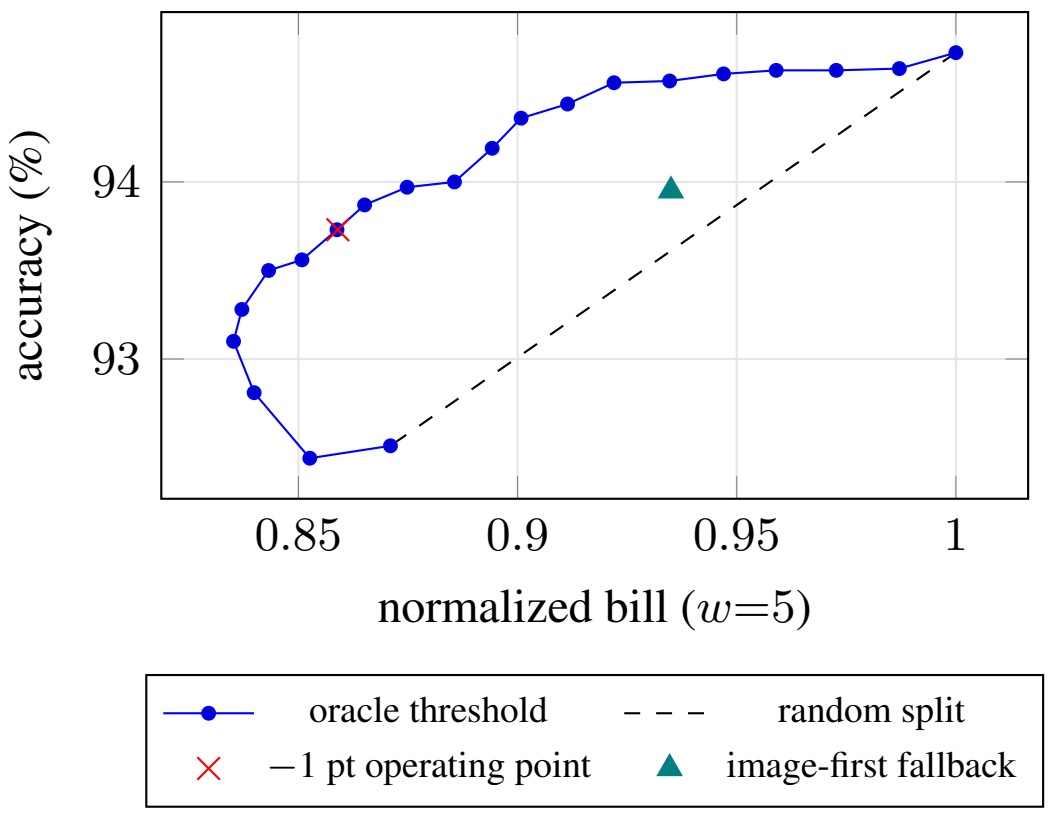
\begin{figure}[t]
\centering
\begin{tikzpicture}
\begin{axis}[width=0.95\columnwidth, height=4.8cm, xlabel={normalized bill ($w{=}5$)}, ylabel={accuracy (\%)}, font=\footnotesize, grid=major, grid style={gray!20}, legend style={font=\scriptsize, at={(0.5,-0.36)}, anchor=north, legend columns=2, /tikz/every even column/.append style={column sep=4pt}}]
\addplot+[mark=*, mark size=1.2pt] table[x=bill, y=acc, col sep=comma] {data/k3_route_curve.csv}; \addlegendentry{oracle threshold}
\addplot+[mark=none, dashed, black] coordinates {(1.0,94.73) (0.871,92.51)}; \addlegendentry{random split}
\addplot+[only marks, mark=x, mark size=3pt, red] coordinates {(0.859,93.73)}; \addlegendentry{$-$1 pt operating point}
\addplot+[only marks, mark=triangle*, mark size=2.5pt, teal] coordinates {(0.935,93.95)}; \addlegendentry{image-first fallback}
\end{axis}
\end{tikzpicture}
\caption{Kimi-K3 on QuALITY (temperature 0.8, 5 samples): accuracy versus bill when the easiest questions by text-mode reasoning length are sent to the image path. The dashed line is a random split. The triangle is the realizable image-first fallback of \S\ref{sec:fallback}.}
\label{fig:route}
\end{figure}

\paragraph{Oracle headroom.} Using the true text-mode reasoning length as difficulty, with the threshold selected and billed on the same samples, routing on Kimi-K3 traces out a frontier that dominates both pure strategies (Figure~\ref{fig:route}). Within a 1-point accuracy budget it reaches a bill of 0.859 (14\% saving); within 0.5 points, 0.901 (ranking questions by two samples and billing on three others: 0.869 and 0.894; on those three samples all-image itself costs 0.862 rather than 0.871, which is why the 0.5-point figure is lower). This rule is a reference point, not an upper bound: a cheaper model's reasoning length does better in gross terms (Table~\ref{tab:pred}). The minimum bill, 0.835, is lower than sending everything to images (0.871) while losing less accuracy (1.6 vs.\ 2.2 points). On LongBench the same threshold rule reaches 0.799 ($-$0.50 points), and a per-question oracle that knows the true $S$ and accuracy difference reaches 0.727 ($-$0.22); for DeepSeek the per-question oracle reaches 0.694 on QuALITY and 0.752 on LongBench. For Qwen and GLM, consistent with \S\ref{sec:breakeven}, difficulty routing saves at most 0.2\% within the same budget, and an oracle that decides from other samples of the same question (selecting questions by their mean saving and accuracy change on the other samples within a 1-point budget, billed on the held-out sample) saves only 3--4\% while losing 0.9--2.1 points: their apparent per-question savings come from which sample happens to run away or fail, not from a property of the question. %

\begin{table}[t]
\centering
\small
\setlength{\tabcolsep}{3pt}
\resizebox{\columnwidth}{!}{%
\begin{tabular}{lcc}
\toprule
Predictor (before generation) & $\rho$ & Bill ($\le$1 pt) \\
\midrule
True length, same samples & 1.00 & 0.859 \\
True length, disjoint samples & 0.79 & 0.869 \\
Random & -- & 0.942 \\
Reasoning router 0.6B (doc+q) & 0.03 & 0.953 \\
Complexity classifier (q only) & 0.14 & 0.975 \\
Document / question length & 0.02 / 0.21 & 0.949 / 0.979 \\
Six further HF routers$^\ast$ & $\le$0.25 & 0.91--0.99 \\
Difficulty rubric, Qwen3.5-9B$^\ast$ & 0.33 & 0.935 \\
Self-reported length, GLM$^\dagger$ & 0.34 & 0.929 \\
Fine-tuned 0.6B on LongBench & AUC 0.51 & $\approx$random \\
Qwen reasoning length & 0.65 & 0.841 / 0.904$^\S$ \\
GLM reasoning length & 0.65 & 0.853 / 0.896$^\S$ \\
\midrule
Image-first fallback (\S\ref{sec:fallback}) & -- & 0.935 \\
\bottomrule
\end{tabular}}
\caption{Routing Kimi-K3 on QuALITY (temperature 0.8; $\rho$ = Spearman correlation with Kimi-K3's text-mode reasoning length). $^\ast$300-question subset, where random routing gives 0.959. $^\dagger$658-question subset, where random routing gives 0.894. $^\S$Gross / net of the router's own cost (Qwen: router priced at 5\% of Kimi-K3; GLM: list prices); the gross figures select and bill on the same samples and are optimistic.}
\label{tab:pred}
\end{table}

\paragraph{Predictors before generation.} We tested every training-free signal we could obtain before the target model generates (Table~\ref{tab:pred}). Off-the-shelf difficulty models---a 0.6B think/no-think router \citep{amirmohseni2025reasoningrouter}, a ModernBERT query-complexity classifier \citep{anasnassar2026llmquerycomplexity}, and six further routers from the Hugging Face Hub---correlate at most 0.25 with Kimi-K3's reasoning length: they score short user prompts for general difficulty, not retrieval difficulty in a long document. Asking a small model for a 10-level difficulty score with explicit criteria reaches a Spearman correlation of 0.33; asking it to state its own output length before answering fails because it does not know (Qwen3.5-9B answers 150 for every QuALITY question; GLM's self-reported length correlates 0.34 with both its own and Kimi-K3's reasoning length on QuALITY). The target model is no better: DeepSeek, asked in a separate non-thinking call on the same images how many tokens it will need, correlates 0.26 with its actual image-mode length on QuALITY and 0.28 within LongBench subsets, with a typical error of 3$\times$. %

\paragraph{A trained predictor learns the task, not the question.} A 0.6B router fine-tuned on Kimi-K3's LongBench outcomes (the 2{,}335 examples with $P_t > 4{,}000$, 5-fold cross-validation grouped by document) reaches an AUC of 0.94 for ``text-mode reasoning exceeds 350 tokens'' and a routed bill of 0.80 at $-$0.6 points. A baseline that ignores the question and outputs the training-fold positive rate of its subset reaches 0.935 and the same bill. Within subsets the out-of-fold AUC is 0.47--0.54 (0.69--0.77 on the training folds), and on QuALITY, an unseen task, 0.51. The LongBench gain is therefore a lookup of the task type: requests carry no dataset label, but their instructions often reveal the task, so routing by task type is a practical option where tasks differ in this way, though it gives no signal within a task. %

\paragraph{The one strong signal costs a second read.} Reasoning length is a stable, largely model-independent property of the question: Kimi-K3's text-mode length correlates 0.79 with itself across disjoint samples and 0.65 with Qwen's and GLM's. Routing on the reasoning length of a cheaper model therefore works in gross terms (bill 0.841--0.853 on QuALITY), but the cheaper model must read the full document, and prefill is 80\% of Kimi-K3's text bill. Net of the router's own cost, the bill is 0.896 (GLM at list prices) and 0.904 (Qwen, priced at 5\% of Kimi-K3): on QuALITY this is lower than the image-first fallback below (0.935), but only by serving a second model that reads every document. %

\paragraph{Information-loss routing with transferred parameters.} The closest prior router, Measure Transport \citep{tang2026visual}, sends an input to the visual path when its transport efficiency $\mathrm{TE}=(1+\gamma-C)\,c_i$ exceeds a margin $\tau$. Here $c_i$ is the input's own ratio of text to visual tokens, and $C=\lambda_1 W+\lambda_2 L\,(1-\mathrm{TRR})$ estimates the information lost in rendering from the answer format ($W$), the BM25 dispersion of question-relevant evidence ($L$) and gzip redundancy (TRR), with $\lambda_1,\lambda_2,\gamma,\tau$ calibrated on synthetic probes. It was evaluated on non-reasoning models, and its routing objective weighs input-token savings against task utility rather than a price-weighted decode bill. We reimplement it with the published parameters (Appendix~\ref{app:details}) and apply it to our paired runs, charging decode at the same price as elsewhere (Table~\ref{tab:mt}). With the 4B operating point it sends 25--79\% of inputs to images. On both LongBench runs it is dominated by fallback, and on Kimi-K3 also by all-image (bill 1.046, above all-text); on QuALITY it moves along the line between all-text and all-image, no better than sending the same fraction at random on DeepSeek (0.846 vs.\ 0.844 at equal accuracy) and slightly better on Kimi-K3 (0.954 vs.\ 0.967). Their diagnostic 32B rule, a bounded-benefit variant that caps the compression benefit at $1+\min(c_i-1,\,0.30)$, sends no input to images and reproduces all-text. With these transferred parameters, the information-loss term is not what separates cheap from expensive inputs: the published implementation assigns one $C$ per dataset, so within QuALITY the rule orders questions by $c_i$ alone, and on Kimi-K3's LongBench a threshold on $c_i$ alone does better than the full rule (0.890 vs.\ 0.930 within a 1-point budget, both thresholds chosen on the evaluation data). Calibrating the rule on our target models might change this; we test only transfer. The compression ratio itself is informative, since it sets the prefill saving of each input; we use it in \S\ref{sec:fallback}. %

\begin{table}[t]
\centering
\footnotesize
\setlength{\tabcolsep}{1pt}
\resizebox{\columnwidth}{!}{%
\begin{tabular}{l@{\hspace{4pt}}cccc}
\toprule
 & \multicolumn{2}{c}{Kimi-K3} & \multicolumn{2}{c}{DeepSeek} \\
\cmidrule(lr){2-3}\cmidrule(lr){4-5}
 & QuALITY & LongBench & QuALITY & LongBench \\
\midrule
All image & 0.871/$-$2.21 & 0.985/$-$0.68 & 0.833/$-$2.60 & 0.864/$-$1.46 \\
MT, 4B & 0.954/$-$0.57 & 1.046/$-$0.80 & 0.846/$-$2.43 & 0.939/$-$1.05 \\
MT, 32B & 1.000/0.00 & 1.000/0.00 & 1.000/0.00 & 1.000/0.00 \\
\midrule
Fallback & 0.936/$-$0.78 & 0.957/$-$0.13 & 0.879/$-$1.20 & 0.896/$-$0.69 \\
\,+ $c_i$ gate & 0.935/$-$0.78 & 0.908/$-$0.27 & 0.879/$-$1.20 & 0.860/$-$0.70 \\
\bottomrule
\end{tabular}}
\caption{Measure Transport (MT) routing \citep{tang2026visual} with its published 4B parameters ($\tau{=}1.28$) and its diagnostic 32B bounded-benefit rule ($\tau{=}1.55$), against image-first fallback (saving budget $T$, $P_0{=}4{,}000$) with and without the gate $c_i \ge 1.60$ on the per-input compression ratio, selected on separate validation sets (\S\ref{sec:fallback}). Bill / accuracy change (points) relative to all-text, prices as in Table~\ref{tab:fallback}. With our dataset-level $C$ the 32B rule's transport efficiency is bounded by $1.30(1{+}\gamma{-}C) < 1.55$, so it routes every input to text by construction and is not a test of MT (Appendix~\ref{app:details}).}
\label{tab:mt}
\end{table}

\subsection{Image-first fallback}
\label{sec:fallback}
If the question cannot tell us in advance, the image path can tell us as it runs. Long image-mode reasoning is the symptom of a question that is hard to read from pixels (\S\ref{sec:tail}), so the image path's own decode length is an online difficulty probe that costs nothing to observe. We therefore send every input with $P_t > P_0$ to the image path and abort to text once image-mode reasoning exceeds $T$, paying for the aborted tokens. Both parameters follow from the break-even accounting of \S\ref{sec:breakeven}. Image-mode decode beyond $T = (P_t - P_v)/w$ has used up the input's prefill saving even if the text path would think for zero tokens, so we use $T$ as a decode budget for the image path. It needs no fitting, but it is not a switching break-even point: it ignores the text path's own reasoning, and whether continuing is cheaper than restarting depends on the remaining image-mode length, which is not known in advance (\S\ref{sec:tail}); a later threshold lets some responses that are about to finish complete. $T$ is proportional to the input length, $T \approx \alpha P_t$ with $\alpha \approx (1-1/c)/w$ (0.087 for Kimi-K3, 0.11--0.12 for DeepSeek). The gate $P_0$ is where the typical question breaks even, $P_0 = w(r-1)\tilde d/(1-1/c) \approx 4{,}000$ tokens for Kimi-K3 on LongBench ($r{=}2.97$, $c{=}1.77$, median $\tilde d{=}175$), matching the empirical crossover of all-image bills between 2--4k (1.06) and 4--16k (0.92--0.97) input tokens on the same run. Both are estimated on test data, and $r$ there is inflated by one subset: with $r{=}2$, $P_0$ would be about 2{,}000. %

\begin{table}[t]
\centering
\small
\setlength{\tabcolsep}{3pt}
\resizebox{\columnwidth}{!}{%
\begin{tabular}{llccc}
\toprule
 & & All image & Saving budget $T$ & $\alpha{=}0.15$ \\
\midrule
\multirow{2}{*}{Kimi-K3} & QuALITY & 0.871 / $-$2.21 & 0.935 / $-$0.78 & 0.889 / $-$0.90 \\
 & LongBench & 0.985 / $-$0.68 & 0.908 / $-$0.27 & 0.911 / $-$0.30 \\
\multirow{2}{*}{DeepSeek} & QuALITY & 0.833 / $-$2.60 & 0.879 / $-$1.20 & 0.868 / $-$1.26 \\
 & LongBench & 0.864 / $-$1.46 & 0.860 / $-$0.70 & 0.860 / $-$0.72 \\
\midrule
GLM 16px & QuALITY & 1.334 / $-$8.42 & 1.004 / $-$0.02 & 1.001 / $-$0.08 \\
Qwen 19px & QuALITY & 1.256 / $-$4.43 & 1.000 / $+$0.00 & 1.000 / $+$0.00 \\
\bottomrule
\end{tabular}}
\caption{Image-first fallback: bill / accuracy change (points) relative to all-text, with the gate $P_t > 4{,}000$ and $c_i \ge 1.60$ (\S\ref{sec:fallback}). Kimi-K3, GLM and Qwen at $w{=}5$, DeepSeek at its list price $w{=}4$. The saving budget $T = (P_t - P_v)/w$ has no tuned parameter. Fallback rates (share of image-path inputs, QuALITY/LongBench): 23/22\% (Kimi-K3), 20/20\% (DeepSeek) for the saving budget $T$; 14/17\% and 17/18\% for $\alpha{=}0.15$. GLM and Qwen inputs almost never pass the gate (median $c_i$ 1.45 and 1.37), so both policies reduce to text. 95\% intervals in Table~\ref{tab:ci}.}
\label{tab:fallback}
\end{table}

\paragraph{Results.} Table~\ref{tab:fallback} shows that fallback turns the all-image accuracy loss of 0.7--2.6 points into 0.3--1.2 points while keeping half or more of the all-image saving on QuALITY and more than all of it on LongBench, where all-image suffers runaway reasoning. The saving budget lies close to the frontier of the $T{=}\alpha P_t$ family, which delays fallback as $\alpha$ grows: for DeepSeek it is within 0.01 of the grid point of equal accuracy (QuALITY 0.879 vs.\ 0.886, LongBench 0.860 vs.\ 0.861); for Kimi-K3 it is the conservative end of the frontier ($\alpha{\approx}0.09$): $\alpha{=}0.15$ lowers the QuALITY bill by a further 0.046 at a cost of 0.12 points, but not the LongBench bill. Delaying further saves at most 5.6 (Kimi-K3) and 2.8 (DeepSeek) more points of the QuALITY bill, mostly by $\alpha{\approx}0.2$, at a cost of only 0.17--0.34 accuracy points, which stays small because the text retry recovers accuracy on the responses that $T$ cuts. On LongBench it saves nothing (DeepSeek 0.858--0.867 up to $\alpha{=}0.3$), and above $\alpha{\approx}0.12$ Kimi-K3's bill rises (0.939 at $\alpha{=}0.3$), since every runaway response pays for the longer wait. Fallback cannot help Qwen and GLM: their inputs almost never pass the compression-ratio gate, so the policy reduces to text (bill 1.000--1.004), and without the gate their best fallback bill within a 1-point budget is 1.20--1.40, because their image-mode accuracy loss is spread over short, non-runaway responses that a length signal does not see. %

\paragraph{Where fallback loses.} Each fallback pays for the image prefill and $T$ wasted decode tokens, together exactly one text prefill ($P_v + wT = P_t$). On DeepSeek's \texttt{passage\_count}, 95\% of image-mode responses fall back, so the subset still costs 1.78$\times$ text (2.35$\times$ without fallback); on long-output tasks such as \texttt{gov\_report}, a threshold proportional to the input cuts off responses that would be long in text mode too. The compression-ratio gate keeps most such inputs on text (15\% of DeepSeek's LongBench \texttt{gov\_report} inputs and 2.5\% and 8.9\% of the full GovReport test set for Kimi-K3 and DeepSeek reach the image path), but those that do mostly fall back, so these sets still cost 1.06, 1.01 and 1.05$\times$ text (1.50, 1.46 and 1.51 without the gate). Where image mode loses no accuracy, fallback only adds cost: on the full Qasper set Kimi-K3's fallback bill is 0.985 against 0.889 for all-image. Fallback beats all-image on cost only where all-image suffers runaway reasoning (LongBench: 0.908 vs.\ 0.985 for Kimi-K3, 0.860 vs.\ 0.864 for DeepSeek); elsewhere what it buys is accuracy. Without the compression-ratio gate, selecting $(P_0, \alpha)$ on one dataset and transferring is stable for Kimi-K3 but not for DeepSeek (selected on LongBench, QuALITY reaches 0.849 at $-$1.68 points), which is why we use $T$, which needs no fitting, as the default; the gates $P_0$ and $t$ are estimated. %

\paragraph{A gate on the per-input compression ratio.} The gate $P_0$ uses the average compression ratio $c$, but each input saves $P_t(1-1/c_i)$ with its own ratio $c_i$, the quantity Measure Transport routes on. On LongBench $c_i$ varies widely (Kimi-K3: 1.42--1.96 between the 10th and 90th percentiles, subset means from 1.37 for \texttt{multi\_news} to 2.09 for \texttt{narrativeqa}), and is associated with the content: it is higher with more line breaks per word, which cost text tokens but add no rendered lines since the renderer collapses all whitespace, and with shorter words; dialogue and transcripts compress best and short documents worst (Spearman correlation of $c_i$ with line breaks per word 0.57, with characters per word $-0.58$, with document length 0.46). Writing the gate as a saving, $P_t(1-1/c_i) > P_0(1-1/c)$, changes the bill by only $-$0.001 to $-$0.007. Additionally requiring $c_i \ge t$ removes the inputs with the smallest saving. We choose $t$ on validation sets disjoint from QuALITY and LongBench, with $P_0{=}4{,}000$ as above; both models select $t{=}1.60$ (Appendix~\ref{app:details}). On the test sets this lowers the LongBench bill from 0.957 (no gate) to 0.908 for Kimi-K3 (95\% interval of the difference $-$0.057 to $-$0.041; accuracy $-$0.14 points, $-$0.29 to $-$0.01) and from 0.896 to 0.860 for DeepSeek ($-$0.044 to $-$0.030; accuracy unchanged), and leaves QuALITY unchanged (Table~\ref{tab:mt}). The excluded inputs are not those with a larger decode multiplier (it is lower: 1.6 vs.\ 3.3--4.4). Their prefill saving is 20--23\% of the text bill against 40--42\% for the rest, and their text-mode outputs are long (about 1{,}000 vs.\ 350--420 tokens): about half are GovReport summaries and over a quarter Qasper. These are the inputs on which a threshold proportional to the input cuts off responses that would be long in text mode too. %

\paragraph{Online signals that do not improve the threshold.} We tried to stop earlier or later than the static threshold using information available during image-mode generation; none helps. A marginal rule that ignores sunk cost and falls back when the expected remaining image decode exceeds the full text cost fails because the expected residual life grows with length (\S\ref{sec:tail}), so the rule is unstable across datasets (DeepSeek on QuALITY: 1.056 when the residual life is estimated on LongBench and 0.916 in-fold, vs.\ 0.879 for the static threshold). Stopping at the $k$-th hesitation word (``wait'', ``hmm'', ``let me re-read'') is no better at any $k$ (Kimi-K3, QuALITY 0.935--1.14 vs.\ 0.935; LongBench, greedy run, 0.94--1.09 vs.\ 0.913): half of all responses contain none, and one or two are normal reasoning. N-gram repetition appears only after the threshold has already fired. Mean token entropy over the first 32--256 tokens predicts exceeding the threshold with an AUC of 0.55--0.61 on QuALITY and 0.41--0.49 within LongBench subsets, and the probability of the end-of-thinking token is near zero until about 15 tokens before the end. This matches the literature: online remaining-length predictors are trained, as probes on hidden states or as fine-tuned self-reports \citep{shahout2024trail,xie2026plp,merzouk2026howmuch,raaschou2025progress}, and the last of these finds that rollouts from the same prefix differ in length, suggesting an empirical floor on prediction error. Inference-time early exit, much of it training-free \citep{yang2025deer,liu2025convergence}, targets a different question, whether the answer is already settled, and could complement the threshold, although only 14--26\% of the extra image-mode tokens follow the final answer (\S\ref{sec:mech}): most of the tax arises before the answer is settled. %

\begin{figure}[t]
\centering
\begin{tikzpicture}
\begin{axis}[width=0.95\columnwidth, height=4.3cm, title={Kimi-K3 ($w{=}5$)}, title style={font=\footnotesize, yshift=-1ex}, ylabel={saving vs.\ text (\%)}, font=\footnotesize, grid=major, grid style={gray!20}, xmin=-3.8, xmax=0.6, ymin=-2, ymax=37, error bars/error bar style={gray!70}]
\addplot+[mark=*, mark size=1.5pt, blue, error bars/.cd, x dir=both, x explicit, y dir=both, y explicit] coordinates {
 (0,0) (0.00,2.54) += (0.40,1.31) -= (0.42,1.23) (-0.24,5.96) += (0.34,1.21) -= (0.38,1.25)};
\addplot+[mark=square*, mark size=1.5pt, red, error bars/.cd, x dir=both, x explicit, y dir=both, y explicit] coordinates {
 (-2.38,12.67) += (0.88,3.06) -= (0.90,3.05) (-2.67,23.59) += (0.92,2.11) -= (0.95,2.16) (-2.68,32.60) += (0.96,1.60) -= (1.02,1.66)};
\addplot+[only marks, mark=triangle*, mark size=2.6pt, teal, error bars/.cd, x dir=both, x explicit, y dir=both, y explicit] coordinates {
 (-0.88,6.47) += (0.51,2.18) -= (0.50,2.49)};
\addplot+[only marks, mark=diamond*, mark size=2.8pt, orange!90!black, error bars/.cd, x dir=both, x explicit, y dir=both, y explicit] coordinates {
 (-0.99,9.72) += (0.62,2.13) -= (0.60,2.41)};
\addplot+[mark=none, dashed, gray] coordinates {(-0.24,5.96) (-2.68,32.60)};
\end{axis}
\end{tikzpicture}

\begin{tikzpicture}
\begin{axis}[width=0.95\columnwidth, height=4.3cm, title={DeepSeek ($w{=}4$)}, title style={font=\footnotesize, yshift=-1ex}, xlabel={accuracy change vs.\ text (points)}, ylabel={saving vs.\ text (\%)}, font=\footnotesize, grid=major, grid style={gray!20}, xmin=-5.6, xmax=0.6, ymin=-2, ymax=43, legend style={font=\scriptsize, at={(0.5,-0.42)}, anchor=north, legend columns=2, /tikz/every even column/.append style={column sep=4pt}}, error bars/error bar style={gray!70}]
\addplot+[mark=*, mark size=1.5pt, blue, error bars/.cd, x dir=both, x explicit, y dir=both, y explicit] coordinates {
 (0,0) (-0.10,4.47) += (0.48,1.11) -= (0.50,1.02) (-0.18,7.04) += (0.51,1.26) -= (0.53,1.16)}; \addlegendentry{text, $B{=}\infty$/1000/300}
\addplot+[mark=square*, mark size=1.5pt, red, error bars/.cd, x dir=both, x explicit, y dir=both, y explicit] coordinates {
 (-2.60,16.71) += (0.80,3.42) -= (0.77,3.62) (-3.82,30.62) += (1.03,2.55) -= (0.96,2.62) (-4.23,38.81) += (1.08,2.00) -= (1.04,2.22)}; \addlegendentry{image, $B{=}\infty$/1000/300}
\addplot+[only marks, mark=triangle*, mark size=2.6pt, teal, error bars/.cd, x dir=both, x explicit, y dir=both, y explicit] coordinates {
 (-1.20,12.08) += (0.55,2.41) -= (0.56,2.68)}; \addlegendentry{image-first fallback}
\addplot+[only marks, mark=diamond*, mark size=2.8pt, orange!90!black, error bars/.cd, x dir=both, x explicit, y dir=both, y explicit] coordinates {
 (-1.23,16.68) += (0.68,2.30) -= (0.64,2.50)}; \addlegendentry{fallback, text $B{=}300$}
\addplot+[mark=none, dashed, gray] coordinates {(-0.18,7.04) (-4.23,38.81)};
\end{axis}
\end{tikzpicture}
\caption{Reasoning budgets (``Keep your thinking within $B$ tokens'') on both paths, all QuALITY questions (3 samples). Saving and accuracy change relative to unbudgeted text; bars are 95\% intervals (articles resampled). The dashed line is a random mix of the two $B{=}300$ paths.}
\label{fig:budget}
\end{figure}
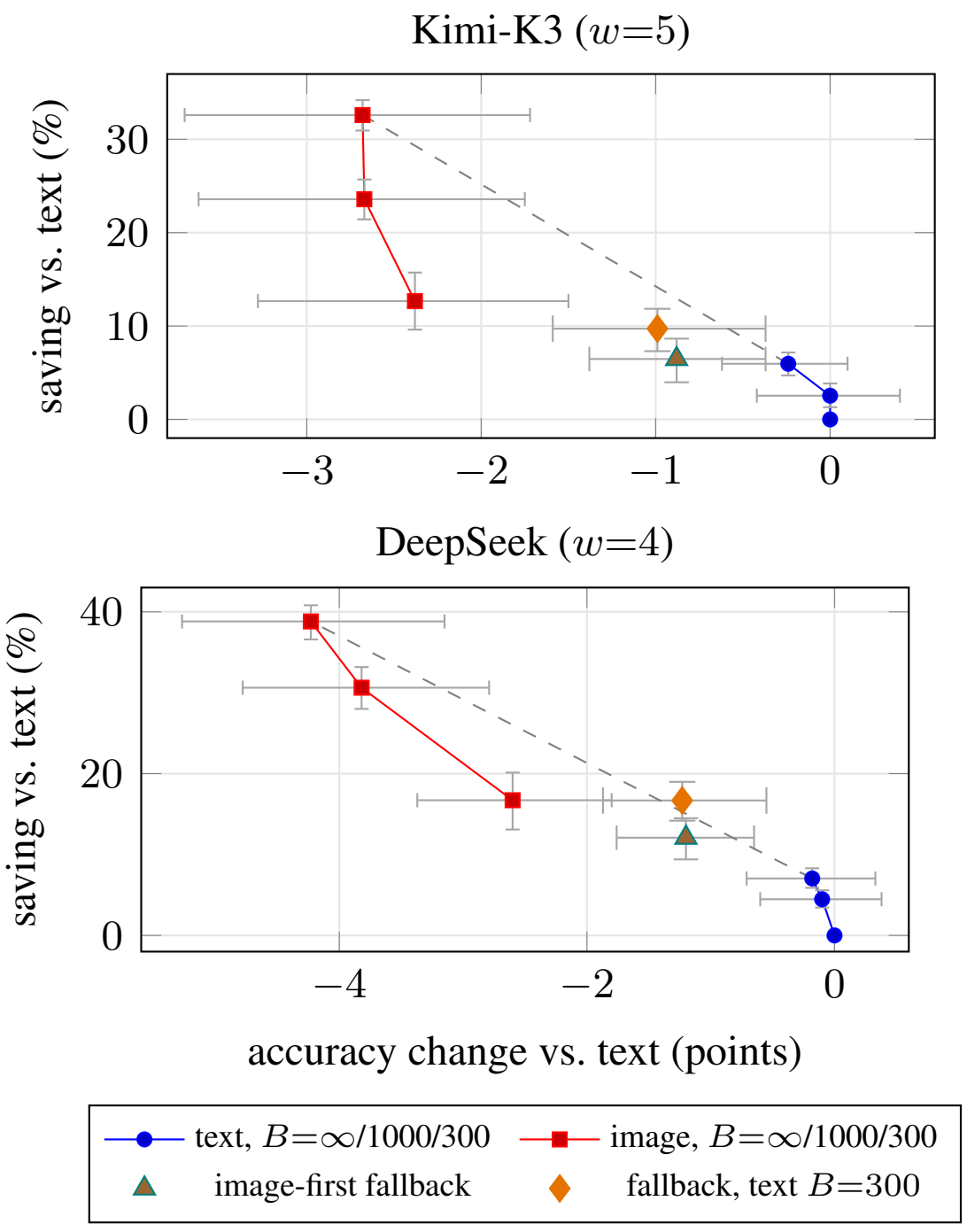

\paragraph{Reasoning budgets on both paths.} A simpler way to shorten reasoning is to ask for it (``Keep your thinking within $B$ tokens'') on both paths. Figure~\ref{fig:budget} shows both models on all 2{,}086 QuALITY questions with 3 samples (for Kimi-K3, the first three of the five in Table~\ref{tab:main}, so its unbudgeted points differ slightly from Table~\ref{tab:fallback}). On the text path the budget is nearly free: $B{=}300$ saves 6.0\% (Kimi-K3) and 7.0\% (DeepSeek) of the bill at $-$0.24 and $-$0.18 points. On the image path it saves more, 23.6\% and 32.6\% at $B{=}1000$ and $B{=}300$ for Kimi-K3 and 30.6\% and 38.8\% for DeepSeek. For Kimi-K3 this costs little beyond the image path's own loss ($-$2.7 points at either budget against $-$2.4 without); for DeepSeek the loss grows from 2.6 to 3.8--4.2 points. In neither model does the budget remove the tax: under the same budget, image-mode reasoning is still 1.78$\times$ and 1.58$\times$ as long as text-mode reasoning for Kimi-K3 (2.09$\times$ without) and 2.24$\times$ and 1.98$\times$ for DeepSeek (2.47$\times$ without). Fallback alone (6.5\% at $-$0.88 for Kimi-K3, 12.1\% at $-$1.20 for DeepSeek) is not on the frontier the budgets span: a random mix of the two $B{=}300$ paths saves as much at $-$0.29 and $-$0.82 points. When fallback's text requests carry the text budget, its saving rises to 9.7\% at $-$0.99 for Kimi-K3 and to 16.7\%, equal to DeepSeek's all-image saving, at $-$1.23 instead of $-$2.60. For DeepSeek this matches the best mix of budgeted paths at the same saving ($-$1.41; paired article-bootstrap difference $+$0.18 points, 95\% CI [$-$0.28, $+$0.63]); for Kimi-K3, whose budgeted image path loses little accuracy, the mix is better ($-$0.58; difference $-$0.41 [$-$0.94, $-$0.02]). On these questions the budget is the stronger lever, and fallback adds to it only where budgeting the image path costs accuracy. %

\begin{table*}[t]
\centering
\scriptsize
\setlength{\tabcolsep}{3pt}
\begin{tabular}{llccccc@{\hspace{6pt}}cccccc}
\toprule
 & & \multicolumn{5}{c}{QuALITY: decode and text path} & \multicolumn{6}{c}{Method} \\
\cmidrule(lr){3-7}\cmidrule(lr){8-13}
 & & & & & & & \multicolumn{2}{c}{QuALITY} & \multicolumn{2}{c}{LongBench} & \multicolumn{2}{c}{Validation (7 sets)} \\
\cmidrule(lr){8-9}\cmidrule(lr){10-11}\cmidrule(lr){12-13}
Model & Effort & $D_t$ & $D_v$ & $D_v/D_t$ & Fallback & Text bill & same & default & same & default & same & default \\
\midrule
\multicolumn{13}{l}{\emph{Full sets, 3 samples}} \\
\multirow{2}{*}{DeepSeek} & 75$^\ast$ & 290 & 716 & 2.48 & 20\% & 1.000 & 0.879/$-$1.20 & 0.879/$-$1.20 & 0.860/$-$0.70 & 0.860/$-$0.70 & 0.825/$-$1.67 & 0.825/$-$1.67 \\
 & 50 & 187 & 566 & 3.04 & 16\% & 0.941 & 0.843/$-$1.50 & 0.793/$-$1.63 & 0.827/$-$0.91 & 0.807/$-$1.36 & 0.822/$-$2.00 & 0.798/$-$1.86 \\
\multirow{2}{*}{Kimi-K3} & max$^\ast$ & 300 & 628 & 2.09 & 23\% & 1.000 & 0.935/$-$0.88 & 0.935/$-$0.88 & 0.908/$-$0.27 & 0.908/$-$0.27 & 0.875/$-$6.22 & 0.875/$-$6.22 \\
 & high & 202 & 308 & 1.52 & 10\% & 0.934 & 0.799/$-$1.53 & 0.746/$-$1.65 & 0.811/$-$0.76 & 0.773/$-$0.89 & 0.820/$-$6.80 & 0.769/$-$7.85 \\
\midrule
\multicolumn{13}{l}{\emph{200-question probes, 1 sample}} \\
\multirow{3}{*}{DeepSeek} & 50 & 156 & 604 & 3.88 & 18\% & 0.910 & 0.865/$-$1.5 & 0.787/$-$4.5 & -- & -- & -- & -- \\
 & 75$^\ast$ & 312 & 720 & 2.31 & 19\% & 1.000 & 0.878/$-$1.5 & 0.878/$-$1.5 & -- & -- & -- & -- \\
 & 100 & 519 & 1210 & 2.33 & 26\% & 1.118 & 0.945/$-$2.5 & 1.057/$-$3.5 & -- & -- & -- & -- \\
\multirow{3}{*}{Kimi-K3} & low & 126 & 193 & 1.53 & 5\% & 0.901 & 0.736/$-$5.0 & 0.663/$-$4.5 & -- & -- & -- & -- \\
 & high & 191 & 355 & 1.86 & 14\% & 0.946 & 0.859/$-$1.5 & 0.813/$-$1.5 & -- & -- & -- & -- \\
 & max$^\ast$ & 267 & 647 & 2.42 & 23\% & 1.000 & 0.927/+0.0 & 0.927/+0.0 & -- & -- & -- & -- \\
\multirow{3}{*}{GLM 16px} & low & 17 & 42 & 2.54 & -- & 0.522 & 0.997/+0.5 & 0.520/$-$2.0 & -- & -- & -- & -- \\
 & high & 63 & 233 & 3.68 & -- & 0.543 & 1.010/+0.0 & 0.548/$-$0.5 & -- & -- & -- & -- \\
 & max$^\ast$ & 1073 & 2136 & 1.99 & -- & 1.000 & 1.005/+0.0 & 1.005/+0.0 & -- & -- & -- & -- \\
\bottomrule
\end{tabular}
\caption{Serving-side reasoning effort ($^\ast$ default). $D_t$, $D_v$: mean text- and image-mode output tokens on QuALITY; fallback: share of image-path inputs; text bill: text path relative to default-effort text. The last six columns give the method (gate $P_t > 4{,}000$, $c_i \ge 1.60$, fallback at $T$) as bill / accuracy change (points), relative to text at the same effort (\emph{same}) and at the default effort (\emph{default}); DeepSeek at $w{=}4$, Kimi-K3 and GLM at $w{=}5$. Kimi-K3 \emph{max} uses the first three of the five QuALITY samples, so it differs slightly from Table~\ref{tab:fallback}. GLM inputs almost never pass the gate (1\%), so its method reduces to text and its fallback rate is not meaningful.}
\label{tab:effort}
\end{table*}

\paragraph{Serving-side reasoning effort.} The chat templates also expose a coarse effort setting (Kimi-K3 and GLM: \emph{low}/\emph{high}/\emph{max}, default \emph{max}; DeepSeek: an integer 1--100 or \emph{low}/\emph{high}/\emph{max}, default \emph{high}, i.e.\ 75, as documented in its official encoding code). The accounting of \S\ref{sec:breakeven} says what to watch: lowering effort cheapens the text path, and it helps the image path only if it shrinks the extra image-mode decode $\Delta D = D_v - D_t$ relative to the saving budget $T$, which effort does not change. On all QuALITY questions (3 samples, the same as in Figure~\ref{fig:budget}; all numbers, including LongBench and the validation sets, in Table~\ref{tab:effort}), Kimi-K3 at \emph{high} halves image-mode reasoning but shortens text-mode reasoning only by a third, so $\Delta D$ falls from 328 to 106 tokens, well below the median $T$ of 627 over image-path inputs, and fallback drops from 23\% to 10\% of image-path inputs. The method's bill relative to same-effort text falls from 0.935 to 0.799 on QuALITY and from 0.908 to 0.811 on LongBench, at 0.5--0.7 more accuracy points lost. DeepSeek at 50 shortens both paths by similar amounts, so its multiplier rises (2.48 to 3.04) but $\Delta D$ falls only from 426 to 379, and the method's bill improves little (QuALITY 0.879 to 0.843); most of its saving against default-effort text comes from the cheaper text path. What decides is the absolute extra decode, not the multiplier. In 200-question probes, the lowest settings cost accuracy (Kimi-K3 \emph{low}: $-$5.0 points for the method), DeepSeek at 100 raises the bill of both paths, and GLM loses 9.5--14 points on the image path at every setting, so its gap lies in reading rather than in reasoning length. The best setting is model-specific and has to be measured; as with budgets, none removes the tax. %

\subsection{Latency}
\label{sec:latency}
We measure end-to-end latency for the two models where image input can pay off, on 20 examples from each of 21 sets: the 14 LongBench subsets, the three full-document sets, and RULER QA at 8k, 32k, 64k and 128k (420 paired requests per model; 418 for Kimi-K3, where two image requests returned server errors). Requests are streamed at concurrency 1, with sampling as in the main runs. We report time to first token (TTFT, including any client-side work) and time per output token (TPOT, decode time divided by $D{-}1$). Fallback latency is assembled from the measured streams: the image request up to its $T$-th token, then the full text request, with the gate and saving budget of Table~\ref{tab:fallback}; it does not include the overhead of cancelling the image request and dispatching the retry.

\paragraph{Server-side rendering.} In a first measurement with the standard pipeline, where the client renders PNG pages and the server decodes and normalizes them, image TTFT was 2.7--2.9$\times$ text TTFT. Profiling attributes most of this to moving pixels rather than to the vision encoder: for Kimi-K3, a 128k-token document becomes 55\,MB of PNG and 785\,MB of normalized patches, which are serialized and broadcast to every tensor-parallel rank. We therefore move rendering into the server. The client lays out the text and sends each page's lines (about 70\,KB per page); the rank that encodes a page rasterizes it on its GPU and applies the model's own resize and normalization. For Kimi-K3 we also enlarge the vision-embedding cache so that chunked prefill does not re-run the encoder. The resulting vision-encoder inputs are bit-identical to those of the PNG pipeline (237 of 237 DeepSeek pages and all tested Kimi-K3 documents), so accuracy is unchanged; client-side layout takes 11--12\,ms per document. All numbers below use this path. %

\begin{table}[t]
\centering
\small
\setlength{\tabcolsep}{3pt}
\resizebox{\columnwidth}{!}{%
\begin{tabular}{llccccc}
\toprule
 & Path & TTFT mean / med (s) & TPOT (ms) & $\bar D$ & Latency mean / p90 (s) & $\times$text \\
\midrule
\multirow{3}{*}{Kimi-K3} & text & 2.77 / 1.53 & 24.9 & 474 & 14.5 / 36.8 & 1.00 \\
 & image & 2.56 / 1.26 & 24.8 & 1{,}160 & 31.4 / 69.1 & 2.16 \\
 & fallback & 2.56 / 1.27 & -- & 678 & 19.6 / 47.7 & 1.35 \\
\midrule
\multirow{3}{*}{DeepSeek} & text & 0.45 / 0.26 & 6.8 & 469 & 3.6 / 9.9 & 1.00 \\
 & image & 0.46 / 0.25 & 6.8 & 1{,}120 & 8.1 / 16.8 & 2.23 \\
 & fallback & 0.45 / 0.25 & -- & 630 & 4.8 / 13.9 & 1.31 \\
\midrule
DeepSeek & image, client PNG & 1.22 / 0.61 & 6.8 & 1{,}097 & 8.7 / 17.2 & 2.31 \\
\bottomrule
\end{tabular}}
\caption{End-to-end latency on 418 (Kimi-K3) and 420 (DeepSeek) requests (21 sets, up to 20 each) at concurrency 1, with server-side rendering. TTFT includes client-side layout; TPOT = median per-request decode time per token; $\bar D$ = mean decode tokens (for fallback, aborted image tokens plus the text response; fallback rows are assembled from the measured streams and exclude cancellation and retry overhead). The last row is the standard client-side PNG pipeline, measured in a separate run against its own text baseline (3.8\,s).}
\label{tab:latency}
\end{table}

\paragraph{Results.} Table~\ref{tab:latency} shows that, once rendering runs on the server, the image path loses no time before the first token and none per token. Image TTFT matches or beats text: 2.56 vs.\ 2.77\,s on Kimi-K3 (faster on 84\% of pairs) and 0.46 vs.\ 0.45\,s on DeepSeek, down from 1.22\,s with client-side PNG. The advantage grows with length, as the shorter image prompt pays off in prefill: at 128k, image TTFT is 17.7 vs.\ 18.2\,s on Kimi-K3 and 2.74 vs.\ 2.89\,s on DeepSeek, while short DeepSeek documents keep a small fixed vision-encoder cost (0.23 vs.\ 0.18\,s at 8k). TPOT is identical on both paths (24.8 vs.\ 24.9\,ms and 6.8\,ms), so at concurrency 1 the shorter image context does not make decode steps measurably faster. The remaining slowdown is therefore the decode tax alone: the image path generates 2.4$\times$ more tokens and is 2.16$\times$ (Kimi-K3) and 2.23$\times$ (DeepSeek) slower end to end. Fallback, which cuts off runaway image-mode reasoning, reduces this to 1.35$\times$ and 1.31$\times$. Image input is still slower than text on 19 of 21 sets for Kimi-K3 and 18 of 21 for DeepSeek, so wherever optical compression lowers the bill, it does so at a cost in latency that the serving optimizations we tested do not remove. %

\section{What Does Not Fix It}

\label{sec:fail}
\begin{table}[t]
\centering
\small
\begin{tabular}{p{2.6cm}p{4.2cm}}
\toprule
Intervention & Outcome \\
\midrule
Lower compression (Qwen) & acc.\ rises monotonically, still $-$4.3 at 1.24$\times$; savings vanish \\
Canvas width 1000--2000\,px & no effect on acc.\ or decode \\
Inline delimiters (``//'') for collapsed breaks & +1.1 acc. \\
Reasoning budget prompt (Qwen) & decode below unbudgeted text, but $-$3.5 acc.\ vs.\ plain image (gives up on hard questions) \\
Locate-then-transcribe & +0 on hard questions vs.\ a fresh-seed baseline, $-$5 on controls, decode $\times$2 \\
Early stop at stable answer & only 14--26\% of the extra tokens follow the final answer (Table~\ref{tab:phase}) \\
Off-the-shelf difficulty routers & at most marginally better than random (\S\ref{sec:route}) \\
Fine-tuned 0.6B router$^\ast$ & learns the task type only (\S\ref{sec:route}) \\
Self-reported length$^\ast$ & $\rho\le0.34$ within a task, error $\approx$3$\times$ \\
Stop at hesitation words$^\ast$ & worse than length fallback at every $k$ \\
Entropy, repetition$^\ast$ & no gain over the static threshold \\
\bottomrule
\end{tabular}
\caption{Interventions on the image path (Qwen unless noted; accuracy changes in points). $^\ast$Routing Kimi-K3; self-reports by GLM, Qwen3.5-9B and DeepSeek, \S\ref{sec:route}.}
\label{tab:fail}
\end{table}
We report interventions on the image path that did not close the gap (Table~\ref{tab:fail}); each is informative about the mechanism.

Two failures are worth spelling out. A reasoning budget does not remove the decode tax: under the same budget, image-mode reasoning stays 1.6--2.2 times as long as text-mode reasoning (\S\ref{sec:fallback}), and for DeepSeek the budget buys its image-path saving with accuracy rather than removing the cause. Prompting the model to first locate and transcribe the relevant passage turns an implicit retrieval problem into an explicit one, yet hard questions do not improve: the model now quotes the right anchor sentence more often but still misses the neighboring sentence or modifier that decides the answer. We also note a methodological trap: selecting ``stable failures'' as questions wrong in all 3 samples at temperature 1.0 admits about one third sampling noise, so any intervention must be compared against a fresh-seed baseline on the same questions. %

\section{Related Work}
\paragraph{Optical context compression.} DeepSeek-OCR \citep{wei2025deepseekocr} and Glyph \citep{cheng2025glyph} established that rendered text can be read back at several-fold compression. Follow-up work adapts the idea to long-context reasoning \citep{wang2026vtcr}, selective re-expansion of relevant pages \citep{xie2026lensvlm,xu2026seer}, adaptive resolution \citep{zhong2026focusvtc}, magnification of selected text spans \citep{zeng2026magnifying}, tables \citep{alonso2026table}, and latent reasoning supervised by rendered chains of thought \citep{wang2026renderofthought,lv2026onelatent}. Evaluations question the gains: DeepSeek-OCR leans heavily on language priors \citep{liang2026visual}, compression fidelity and task accuracy diverge \citep{gao2026zerosense}, most evaluated VLMs struggle with long dependencies \citep{zhao2025vtcbench}, information per visual token has sharp limits \citep{zhuang2026information}, and among small encoders trained from scratch behind a shared bottleneck no encoding of tokens, bytes or pixels wins across tasks \citep{ziegler2026rateutility}. Measure Transport \citep{tang2026visual} decides when to take the visual path from estimated information loss and input savings, evaluated on non-reasoning models; we test its rule once decode is billed and build on its compression-ratio signal (\S\ref{sec:route}, \S\ref{sec:fallback}). On the decode side, downsampled tables lengthen reasoning \citep{alonso2026table} and commercial APIs emitted more thinking tokens for rendered context in a practitioner test \citep{boluk2026snapcompact}; we measure the effect under controlled long-document conditions across five models and relate it to the bill. Concurrent work finds that VLMs recognize rendered text more reliably than they use it, and that verbalizing the evidence, for instance through the model's own transcript, closes much of the gap \citep{shi2026aloud}; our self-transcription round trip (\S\ref{sec:diag}) is a diagnostic in the same spirit, while our focus is the decode cost and when it cancels the prefill saving.

\paragraph{Modality gap in rendered text.} Matched content presented as text or pixels can yield different answers \citep{vansprang2025same,pottanigari2026text,sun2026reading}. Mechanistic studies identify sparse retrieval heads that locate evidence in images and rendered text \citep{li2026retrieval,pei2026vera}, and that models localize evidence without using it \citep{zeng2026magnifying}. Agentic settings show systematic policy drift when histories are read as images \citep{fan2026reading}. We connect these findings to reasoning length and cost.

\paragraph{Compression and output length.} Lossy inputs lengthen outputs in other compression families too: prompt compression \citep{adeyemi2026cavewoman}, though \citet{johnson2026prompt} see it only at the most aggressive ratio, by 3\% and not significantly, and KV-cache compression \citep{liu2025hold,zhu2026tokens}. Generic reasoning-shortening methods reduce length, though accuracy can drop for distilled models, increasingly on harder tasks \citep{wang2025wait}. Routing between models on the predicted chance that the stronger model wins \citep{ong2024routellm}, output-length prediction for scheduling \citep{jin2023s3,zheng2023response,tao2025ranking}, and predicting accuracy as a function of reasoning budget for token allocation \citep{brown2026predictive} are established, and reasoning length is linearly predictable from question activations \citep{sheng2025reasoning}. Image-first fallback is a cascade in the sense of \citet{dekoninck2024unified}, but between two input forms of the same model, and it escalates on the image path's own decode length rather than on a quality estimate. Online remaining-length predictors are discussed in \S\ref{sec:fallback}; for large untuned models, asking for the length at the start of the same response is more accurate than asking for it alone \citep{zheng2023response}. Reasoning models overthink easy questions \citep{chen2024overthinking,sui2025stop}. Their reasoning length can be controlled by training \citep{aggarwal2025l1}, by a budget stated in the prompt, which small budgets fail to enforce \citep{han2024tale}, by budget forcing \citep{muennighoff2025s1}, or by stopping once a trial answer is confident or stable \citep{yang2025deer,liu2025convergence}; at higher provider effort levels, accuracy gains diminish or vanish while tokens keep growing \citep{srivastava2025overthink,wani2026reasoningtax}. These studies vary reasoning length for a fixed text input; we vary the input modality and find that budgets and effort settings shorten both paths but leave the image path's extra reasoning in place (\S\ref{sec:fallback}). We choose between \emph{input modalities} of the same model, and show that the question-level signal needed to do so before generation is not available from the training-free signals we tested, while the length of the image path's own reasoning is.

\section{Conclusion}
Optical context compression saves input tokens but, on reasoning VLMs, makes models think longer. The model can still read the text, since its own transcription restores text-mode reasoning, and most of the extra reasoning occurs before the final answer is fixed; whether it goes to retrieving evidence, integrating it or a changed reasoning strategy, our diagnostics do not separate. The inflation lengthens reasoning by a roughly constant factor for Kimi-K3 and by a factor that shrinks with difficulty for Qwen and GLM. An exchange-rate decomposition---relative input saving per relative output increase against the workload's price-weighted output-to-input ratio---explains when compression pays off; among the four general-purpose models it holds for Kimi-K3 and DeepSeek and fails for Qwen and GLM, mainly because their verbose text-mode reasoning raises the bar; Glyph clears it on cost but loses accuracy. Per-question routing has large oracle headroom, but none of the training-free signals we tested captures it before generation; an image-first policy that falls back to text once image-mode reasoning has used up the input's prefill saving captures part of it with no learned component, and gating it on each input's compression ratio, the signal of the closest prior router, extends the saving on LongBench; on QuALITY, reasoning budgets on both paths are the stronger lever. The tax also shows up as latency: with rendering moved to the server, time to first token and time per output token match text, yet image input is 2.2$\times$ slower end to end, entirely through longer decoding, and fallback still leaves it 1.3--1.4$\times$ slower. Evaluations of optical compression should report decode alongside prefill, and its heavy tail alongside its mean.

\section*{Limitations}
\emph{Scope and data.} Our analysis covers five models on QuALITY and LongBench, and two of them on four further long-context sets; tasks are multiple-choice, short-answer QA and summarization, and the decode multiplier may differ for agentic tasks. RULER is synthetic and we use only its single-hop QA task (qa\_1), with 100 examples per length. Rendering excludes non-Latin scripts and code, and it collapses all whitespace, including paragraph breaks, while the text arm keeps it: part of the compression ratio comes from dropped formatting, and the image path loses structure that inline delimiters partly restore (Table~\ref{tab:fail}); we did not run a text arm with collapsed whitespace. Rendering parameters were chosen on QuALITY itself, and on LongBench the image prompts differ from the text prompts in some instruction wording (\S\ref{sec:setup}). DeepSeek's 19 longest \texttt{narrativeqa} examples are excluded rather than rerun with the larger limit used later, and apart from the reported greedy runs Kimi-K3 was sampled only at temperature 0.8. LongBench runs use 3 samples per example (one for the DeepSeek 28\,px control), so per-subset differences of about 1 point are within noise; the decode-form and predictor analyses of Kimi-K3 on LongBench use the earlier greedy run.

\emph{Cost and latency.} The bill model uses a single price ratio, ignores long-context price tiers and prefix caching, and does not capture the lower per-step decode compute of shorter contexts. The 16k generation limit truncates the heaviest image-mode tails (up to 26\% on \texttt{passage\_count}, where no text-mode response is truncated, so $r$ and the normalized bill there are lower bounds). Latency is measured at concurrency 1 on 418--420 requests per model with server-side rendering; under batching, the shorter image context may lower per-step decode cost, which we did not measure (\S\ref{sec:latency}).

\emph{Method selection.} The fallback results use the same data to report and, for $\alpha{=}0.15$, to select the threshold; the saving budget $T$ has no tuned parameter, but its gate $P_0$ was estimated on Kimi-K3's LongBench run, and the compression-ratio gate was selected on validation sets for the same two models only. Non-default reasoning efforts were run on the full sets for one setting per model (DeepSeek 50, Kimi-K3 \emph{high}) and otherwise only on 200-question probes, with the gates kept at their default-effort values. Our reimplementation of Measure Transport uses its published parameters, calibrated on Qwen3-VL, and fills in details the paper leaves open (Appendix~\ref{app:details}); we did not rerun its probes on our models, so the comparison measures transfer, not a rule calibrated for them. We excluded hidden-state probes by design, since they require the target model's prefill; a trained online length predictor might improve on the static threshold. More generally, our negative results on prediction cover the signals we tested and do not show that the tax is unpredictable in principle.

\emph{Analysis.} Commitment points in reasoning traces are detected with rules refined on about 17 traces and found in 59--69\% of DeepSeek's traces; they were checked against an LLM annotator (the same model used as an assistant) on 100 traces of up to 20{,}000 characters, without human annotation, which agrees on the share before the final answer but places the first commitment earlier; token counts per phase are approximated by character fractions. The mechanistic account in \S\ref{sec:mech} is a working model consistent with our measurements, not a causal proof; ``pre-commitment'' tokens are a positional category and need not all be evidence search. Longer reasoning under compressed input has been reported for tables \citep{alonso2026table} and informally for commercial APIs \citep{boluk2026snapcompact}; our contribution is its controlled characterization on long documents and the resulting decision rule.

\section*{Use of AI Assistants}
The author designed and directed this research and checked the results, the references and the text. Claude Opus~5.5 (Anthropic) was used as an assistant to write and run code and to draft and edit the text.

\bibliography{refs}

\appendix
\renewcommand{\dbltopfraction}{0.9}
\renewcommand{\textfraction}{0.05}
\section{Experimental Details}
\label{app:details}\label{app:tables}

\paragraph{Rendering.} Documents are typeset in Times New Roman, black on white, with a 28\,px margin and a line height of 1.18 times the font size, drawn at 3$\times$ resolution and downsampled with a Lanczos filter for anti-aliasing. All whitespace, including paragraph breaks, is collapsed to single spaces. Table~\ref{tab:render} lists the per-model settings. For QuALITY, each article is a single tall image within each model's pixel limits, so no processor downscales it (DeepSeek: paginated as for LongBench), and the prompt replaces the article by ``The text is shown in the images''; for LongBench, documents are paginated so that no page is downscaled by the processor (Kimi-K3: at most 400 lines, about 6.5k\,px, per page, below its 512-patch side limit; DeepSeek: pages of at most 1{,}134\,px cut between text lines, about 1{,}000 visual tokens each, below its 1{,}024-token per-image limit). Pages are placed before the question in a single user message. Glyph uses its released rendering configuration and pipeline. %

\begin{table}[h]
\centering
\small
\setlength{\tabcolsep}{4pt}
\begin{tabular}{lccc}
\toprule
Model & Font (px) & Width (px) & Comp. \\
\midrule
Qwen & 19/14 & 1750/1500 & 1.36/2.41$\times$ \\
GLM & 18/16 & 1500 & 1.13/1.43$\times$ \\
Kimi-K3 & 14 & 1500 & 1.78$\times$ \\
DeepSeek & 20/28 & 1512 & 1.97/1.03$\times$ \\
Glyph & \multicolumn{2}{c}{official} & 1.79$\times$ \\
\bottomrule
\end{tabular}
\caption{Rendering settings. Comp.\ = prefill compression on QuALITY; $a/b$ = the two settings reported in Table~\ref{tab:main} (DeepSeek 28\,px is the no-compression control). Font sizes were swept per model on QuALITY (Qwen 14--20\,px, canvas width 1000--2000\,px; GLM 16--28\,px; DeepSeek 16--28\,px).}
\label{tab:render}
\end{table}

\paragraph{Serving and sampling.} Kimi-K3 is served with SGLang, tensor-parallel over 16 H200 GPUs on two nodes; DeepSeek with vLLM 0.32 (one tensor-parallel replica of 8 GPUs per node); Qwen with SGLang (8 data-parallel replicas per node); GLM with vLLM. Sampling uses each model's recommended or reported thinking-mode setting: Qwen temperature 1.0, top-$p$ 0.95, top-$k$ 20; GLM temperature 1.0, top-$p$ 0.95; DeepSeek temperature 1.0, top-$p$ 1.0 with its default reasoning effort (75 of 100); GLM and Kimi-K3 at their default, highest effort (\emph{max}); Kimi-K3 temperature 0.8, a deviation from its reported evaluation setting (temperature 1.0, top-$p$ 0.95) that we keep identical on both benchmarks. Each sample uses a fixed seed ($1000k{+}7$ for sample $k$). QuALITY uses 3 samples per question (Kimi-K3: 5, plus one greedy run), LongBench 3 samples per example (Kimi-K3 also has one earlier greedy run, used only for the decode-form and predictor analyses; the DeepSeek 28\,px control has one sample). The generation limit is 16{,}000 tokens; responses that reach it are scored as wrong (QuALITY) or scored on whatever answer text they contain (LongBench, usually empty). On QuALITY at most 0.5\% of responses reach the limit in either mode, except GLM's image path (2.6\% at 18\,px, 4.3\% at 16\,px); on LongBench 1.9\% (Kimi-K3) and 1.2\% (DeepSeek) of image-mode and at most 0.04\% of text-mode responses do, mostly on \texttt{passage\_count} (image mode: 25.7\% and 14.7\%). Text and image arms of the same question and sample index are paired in every comparison.

\paragraph{Prompts.} QuALITY: the article, a blank line, then ``Question: \ldots'', the four options as ``A.''--``D.'' lines, and ``Answer with just the letter (A/B/C/D).'' The image arm replaces the article with the rendered image(s) and keeps the rest verbatim. LongBench: the text arm uses the official per-subset prompts; the image arm uses the image prompts released with Glyph, which replace the context by a reference to the images and also reword some instructions (\S\ref{sec:setup}).

\paragraph{Answer parsing (QuALITY).} The final answer (after the reasoning) is mapped to a letter by, in order: a letter at the start of the answer after stripping whitespace and Markdown characters; otherwise the last ``answer is X'' / ``Answer: X'' pattern; otherwise the last line consisting of a single letter. Kimi-K3 answers often begin with Markdown emphasis, so for it only the first rule is applied after stripping whitespace, asterisks, underscores, backticks, hashes, angle brackets and brackets. Unparseable answers count as wrong.

\paragraph{Commitment detection (Table~\ref{tab:phase}).} A sentence commits to a candidate if it contains a conclusion cue (e.g., ``answer'', ``choose'', ``go with'', ``so'', ``therefore'', ``must be'', ``final'') followed within 40 characters by a capital option letter, or an option letter followed by ``is correct/best'' or ``fits/matches'', excluding negated mentions (``B is wrong'', ``not C'') and restatements of the option list. The first commitment ends the pre-commitment phase; the last change of the committed letter ends the oscillation phase; the remainder is verification. The rules were refined by manual inspection of about 17 traces (fixing, e.g., the article ``a'' read as option A and option letters inside words). Phase lengths are measured in characters and converted to tokens in proportion to the trace's total token count. %

\paragraph{Validation against LLM annotation.}\label{app:phasevalid} To check the rules beyond the traces used to refine them, an LLM annotator (Claude Opus~5.5, the assistant model named in the AI statement, run separately and blind to the rule output and to the input modality, and shown only the numbered sentences of each trace) marked the first commitment and the point after which the answer no longer changes on 100 traces: 10 randomly sampled both-correct pairs per model (Qwen 19px, GLM 18px, Glyph, Kimi-K3, DeepSeek 20px), excluding pairs with a trace longer than 20{,}000 characters. The median absolute disagreement is 5\% of the trace length for the first commitment and 3\% for the final-answer point (below 1\% for Kimi-K3 and DeepSeek, 8--13\% for Qwen, GLM and Glyph). On these 50 pairs the rule and the annotator both place 73\% of the extra decode before the final answer is fixed and attribute 26\% and 27\% to verification. The annotator places the first commitment earlier, so it splits the part before the final answer into 41\% pre-commitment and 32\% oscillation, against 53\% and 20\% for the rule. We therefore treat the share before the final answer as the robust quantity and the boundary between pre-commitment and oscillation as approximate. %

\begin{table*}[t]
\centering
\small
\setlength{\tabcolsep}{4pt}
\resizebox{\textwidth}{!}{%
\begin{tabular}{llccc}
\toprule
Model & Data & All image & Saving budget $T$ & $\alpha{=}0.15$ \\
\midrule
Qwen 19px & QuALITY & 1.256 [1.207, 1.310] / $-$4.43 [$-$5.61, $-$3.20] & 1.000 [1.000, 1.000] / $+$0.00 [$+$0.00, $+$0.00] & 1.000 [1.000, 1.000] / $+$0.00 [$+$0.00, $+$0.00] \\
GLM 16px & QuALITY & 1.334 [1.250, 1.425] / $-$8.42 [$-$9.78, $-$7.06] & 1.004 [1.000, 1.010] / $-$0.02 [$-$0.05, $+$0.00] & 1.001 [1.000, 1.004] / $-$0.08 [$-$0.21, $+$0.00] \\
Kimi-K3 & QuALITY & 0.871 [0.841, 0.902] / $-$2.21 [$-$2.97, $-$1.46] & 0.935 [0.914, 0.957] / $-$0.78 [$-$1.27, $-$0.31] & 0.889 [0.866, 0.912] / $-$0.90 [$-$1.46, $-$0.36] \\
Kimi-K3 & LongBench & 0.985 [0.964, 1.005] / $-$0.68 [$-$1.16, $-$0.15] & 0.908 [0.895, 0.922] / $-$0.27 [$-$0.59, $+$0.06] & 0.911 [0.897, 0.925] / $-$0.30 [$-$0.66, $+$0.07] \\
DeepSeek ($w{=}4$) & QuALITY & 0.833 [0.799, 0.869] / $-$2.60 [$-$3.44, $-$1.75] & 0.879 [0.853, 0.906] / $-$1.20 [$-$1.78, $-$0.66] & 0.868 [0.843, 0.896] / $-$1.26 [$-$1.90, $-$0.70] \\
DeepSeek ($w{=}4$) & LongBench & 0.864 [0.849, 0.879] / $-$1.46 [$-$1.94, $-$1.00] & 0.860 [0.845, 0.874] / $-$0.70 [$-$1.05, $-$0.35] & 0.860 [0.847, 0.873] / $-$0.72 [$-$1.04, $-$0.39] \\
\bottomrule
\end{tabular}}
\caption{Bootstrap 95\% intervals (2{,}000 paired resamples, percentile) for bill / accuracy change in points, with the gate $P_t > 4{,}000$ and $c_i \ge 1.60$ for the fallback columns as in Table~\ref{tab:fallback}. Each resampled unit carries all its samples; QuALITY resamples whole articles (questions on one article are correlated), LongBench resamples examples within each subset (macro-average). At $w{=}5$, all-image DeepSeek bills are 0.898 [0.860, 0.940] (QuALITY) and 0.929 [0.912, 0.947] (LongBench).}
\label{tab:ci}
\end{table*}

\begin{table*}[t]
\centering
\scriptsize
\setlength{\tabcolsep}{3pt}
\begin{tabular}{lll@{\hspace{6pt}}cccccccc}
\toprule
 & & & \multicolumn{8}{c}{$t$} \\
\cmidrule(lr){4-11}
 & & & 1.00 & 1.50 & 1.55 & 1.60 & 1.65 & 1.70 & 1.80 & 2.00 \\
\midrule
\multirow{2}{*}{Validation} & Kimi-K3 & 7 sets & 1.017/$-$6.62 & 0.893/$-$6.67 & 0.874/$-$6.69 & \textbf{0.875/$-$6.22} & 1.002/$-$0.10 & 1.000/$-$0.11 & 0.996/$-$0.05 & 0.994/$-$0.04 \\
 & DeepSeek & 7 sets & 0.954/$-$1.69 & 0.874/$-$1.68 & 0.838/$-$1.69 & \textbf{0.825/$-$1.67} & 0.979/$-$0.30 & 0.994/+0.04 & 0.998/+0.02 & 0.996/+0.01 \\
\midrule
\multirow{4}{*}{Test} & \multirow{2}{*}{Kimi-K3} & QuALITY & 0.936/$-$0.78 & 0.935/$-$0.78 & 0.935/$-$0.78 & \textbf{0.935/$-$0.78} & 0.935/$-$0.78 & 0.935/$-$0.78 & 0.946/$-$0.56 & 1.000/$-$0.03 \\
 & & LongBench & 0.957/$-$0.13 & 0.919/$-$0.21 & 0.912/$-$0.27 & \textbf{0.908/$-$0.27} & 0.908/$-$0.37 & 0.905/$-$0.55 & 0.904/$-$0.42 & 0.933/$-$0.21 \\
 & \multirow{2}{*}{DeepSeek} & QuALITY & 0.879/$-$1.20 & 0.879/$-$1.20 & 0.879/$-$1.20 & \textbf{0.879/$-$1.20} & 0.879/$-$1.20 & 0.879/$-$1.20 & 0.883/$-$1.20 & 0.936/$-$0.74 \\
 & & LongBench & 0.896/$-$0.69 & 0.873/$-$0.67 & 0.863/$-$0.68 & \textbf{0.860/$-$0.70} & 0.855/$-$0.68 & 0.861/$-$0.61 & 0.878/$-$0.46 & 0.925/$-$0.28 \\
\bottomrule
\end{tabular}
\caption{Selection of the compression-ratio gate: an input goes to the image path iff $P_t > 4{,}000$ and $c_i \ge t$, then falls back at the saving budget $T$; $t{=}1.00$ is the ungated policy. Bill / accuracy change (points) relative to all-text; validation accuracy is macro-averaged over the seven validation sets. $t{=}1.60$ (bold) is selected on the validation rows only (Appendix~\ref{app:details}); the test rows are reported, not used for selection.}
\label{tab:gateabl}
\end{table*}

\paragraph{LongBench.} We use the 14 English, non-code subsets with their official metrics: F1 for narrativeqa, qasper, multifieldqa\_en, hotpotqa, 2wikimqa, musique and triviaqa; ROUGE-L for gov\_report, qmsum, multi\_news and samsum; classification accuracy for trec; exact counting accuracy for passage\_count; retrieval accuracy for passage\_retrieval\_en. The overall score is the unweighted mean over subsets. Inputs are not truncated; for DeepSeek, whose server accepts at most 65{,}536 tokens, the 19 narrativeqa examples whose text prompt exceeds this limit are excluded from both arms (2{,}731 of 2{,}750 pairs). %

\paragraph{Full-size sets and RULER.} Qasper: all 1{,}451 test questions (v0.3); the context is the paper body as in LongBench (section titles and paragraphs, no abstract), and each annotator contributes one reference (``unanswerable'', ``yes''/``no'', the extractive spans joined by commas, or the free-form answer). GovReport: all 973 test reports with their reference summaries. QMSum: all general and specific queries of the 35 test meetings, with the transcript as ``speaker: utterance'' lines. These use the LongBench prompts and metrics of the corresponding subsets (F1, ROUGE-L). RULER: the official generator for task qa\_1 (SQuAD) with seed 42 and the Kimi-K3 tokenizer, 100 examples per maximum length (8{,}192 to 131{,}072 tokens including the 32-token answer budget); the text arm uses the official template, the image arm replaces the documents by ``The documents are shown in the images'', and scoring uses RULER's substring-match metric. Contexts above 128k Kimi-K3 tokens would be middle-truncated as in LongBench; no example of these three sets reaches this length (maximum 39.8k), and RULER is not truncated. DeepSeek is served with a maximum length of 196{,}608 tokens for these runs; a 128k RULER document becomes 7 images for Kimi-K3 and 79 for DeepSeek. %

\paragraph{Cost and fallback accounting.} Prefill and decode token counts are those reported by the serving engine (image tokens included in prefill). For image-first fallback, an aborted request is charged its full image prefill plus $T$ decode tokens, followed by a complete text request; the reported accuracy is that of the text answer. The prefill gate $P_0{=}4{,}000$ is applied on the text token count. The compression ratio $c_i = P_t/P_v$ uses the full prompt token counts of both arms; it needs only the page count, which the client's layout step yields without rendering.

\paragraph{Measure Transport reimplementation.} We follow Eqs.~8--12 of \citet{tang2026visual}: the visual path is taken iff $(1+\gamma-C)\,c_i \ge \tau$ with $C=\lambda_1 W+\lambda_2 L(1-\mathrm{TRR})$, using their calibrated $(\lambda_1,\lambda_2,\gamma,\tau)=(0.213, 0.627, 0.069, 1.28)$ for 4B and $(0.053, 0.233, 0.241, 1.55)$ for 32B. The 32B rule is their diagnostic bounded-benefit variant (their App.~F.2), not their main rule: it replaces $c_i$ by $1+\min(c_i-1,\,0.30)$, so $\mathrm{TE}\le1.30\,(1+\gamma-C)$, which stays below $\tau{=}1.55$ for all our dataset-level $C$ (with per-input $C$, at most 3\% of inputs reach the image path). Where the paper does not specify a detail we choose as follows. $W$ takes the tier values of their precision probe: 0.10 for coarse classification (\texttt{trec}) and 0.65 for all other formats, which are option-level, extractive, numerical or generative. $L$ is the normalized entropy of BM25 scores ($k_1{=}1.5$, $b{=}0.75$) between the question (with options, for QuALITY) and 64-word segments of the context, set to 1 when no segment matches. $\mathrm{TRR} = 1 - |\mathrm{gzip}(x)|/|x|$ in bytes. As in their implementation, $L$ and TRR are averaged per dataset (per LongBench subset), so $C$ is constant within a dataset; computing $C$ per input changes bills by at most 0.045 and does not change which policies dominate. Because $C$ is constant on QuALITY, these choices affect only the overall fraction routed to images there, which is set by $\tau$ (Table~\ref{tab:mt} uses the published 4B and 32B values). %

\paragraph{Selecting the compression-ratio gate.} The validation sets are the full Qasper, GovReport and QMSum test sets and RULER QA at 8k, 32k, 64k and 128k, with the three seeds of the Kimi-K3 and DeepSeek runs (9{,}300 paired answers per model). We checked them against the test sets for shared documents: no validation document shares a prefix with a test document, and by random 80-character substrings 5 GovReport documents overlap with LongBench; we remove them. $P_0$ is fixed at 4{,}000 (\S\ref{sec:fallback}) and $t$ is scanned from 1.00 to 2.30 in steps of 0.05. We select the lowest validation bill whose accuracy change (macro-averaged over the seven sets) is no lower than that of the ungated policy ($t{=}1$). A fixed accuracy budget does not work on these sets: all-image Kimi-K3 loses 7.65 points there, so a one-point budget forces nearly every input to text. Each model alone and the two jointly select $t{=}1.60$ (Table~\ref{tab:gateabl}). On the validation sets this value lies at an edge: their $c_i$ is lower than on the test sets (median 1.48 for Kimi-K3, 1.57 for DeepSeek), and from $t{=}1.65$ on almost no input goes to images, so the bill returns to 0.98--1.00. The selected policy loses 6.22 points on these sets, about as much as the ungated one (6.62), so the gate does not add this loss; note that the macro-average gives RULER four of the seven sets. On the test sets the LongBench bill varies by at most 0.03 between 1.55 and 1.80 (0.904--0.912 for Kimi-K3, 0.855--0.878 for DeepSeek), so the selected value is not at an edge there. %

\end{document}